\documentclass[11pt]{article}

\usepackage[]{acl}

\usepackage{times}
\usepackage{latexsym}

\usepackage[T1]{fontenc}

\usepackage[utf8]{inputenc}

\usepackage{microtype}

\usepackage{xspace,mfirstuc,tabulary}
\usepackage{booktabs}
\usepackage{amssymb}
\usepackage{amsmath}
\usepackage{multirow,booktabs, hhline}
\usepackage[ruled,noend]{algorithm2e}
\usepackage{amsmath, bm}

\usepackage{graphicx}
\usepackage{color}
\usepackage{bbm}
\usepackage{bbding}
\usepackage{subfigure}
\usepackage{cleveref}
\crefname{section}{§}{§§}
\Crefname{section}{§}{§§}

\DeclareMathOperator*{\argmin}{arg\,min}

\usepackage{enumitem}

\newenvironment{itemize*}%
 {\leftmargini=20pt\begin{itemize}%
  \setlength{\itemsep}{3pt}%
  \setlength{\parskip}{0pt}%
  }%
 {\end{itemize}}
\newenvironment{enumerate*}%
 {\begin{enumerate}%
  \setlength{\itemsep}{0pt}%
  \setlength{\parskip}{0pt}}%
 {\end{enumerate}}

\title{Recyclable Tuning for Continual Pre-training}

\author{
 Yujia~Qin$^{1}\thanks{\ \ Indicates equal contribution.}$\hspace{0.5em}, Cheng~Qian$^{1*}$, Xu~Han$^{1\dag}$, Yankai~Lin$^{2}$, Huadong~Wang$^{1}$, \textbf{Ruobing Xie}$^3$, \\ \textbf{Zhiyuan Liu}$^{1}\thanks{\ \  Corresponding author.}$\hspace{0.4em}, \textbf{Maosong Sun}$^{1\dag}$, \textbf{Jie Zhou}$^{3}$ \\
 $^1$NLP Group, DCST, IAI, BNRIST, Tsinghua University, Beijing \\
 $^2$Gaoling School of Artificial Intelligence, Renmin University of China, Beijing \\
 $^3$Pattern Recognition Center, WeChat AI, Tencent Inc. \\
\texttt{\{qyj20, qianc20\}@mails.tsinghua.edu.cn}\\
}

\begin{document}
\maketitle

\begin{abstract}
Continual pre-training is the paradigm where pre-trained language models (PLMs) continually acquire fresh knowledge from growing data and gradually get upgraded. Before an upgraded PLM is released, we may have tuned the original PLM for various tasks and stored the adapted weights. However, when tuning the upgraded PLM, these outdated adapted weights will typically be ignored and discarded, causing a potential waste of resources. We bring this issue to the forefront and contend that proper algorithms for recycling outdated adapted weights should be developed. To this end, we formulate the task of recyclable tuning for continual pre-training. In pilot studies, we find that after continual pre-training, the upgraded PLM remains compatible with the outdated adapted weights to some extent. Motivated by this finding, we analyze the connection between continually pre-trained PLMs from two novel aspects, i.e., mode connectivity, and functional similarity. Based on the corresponding findings, we propose both an initialization-based method and a distillation-based method for our task. We demonstrate their feasibility in improving the convergence and performance for tuning the upgraded PLM. We also show that both methods can be combined to achieve better performance. The source codes are publicly available at \url{https://github.com/thunlp/RecyclableTuning}.
\end{abstract}
\section{Introduction}
The emergence of pre-trained language models (PLMs) has revolutionized the entire field of natural language processing (NLP)~\citep{bommasani2021opportunities}. Through downstream adaptation, PLMs effectively stimulate the knowledge acquired during pre-training and achieve remarkable success in various downstream tasks~\citep{devlin2018bert,liu2019roberta,2020t5}.
Such adaptation can be achieved by either full-parameter fine-tuning or parameter-efficient tuning~\citep{pmlr-v97-houlsby19a}, and the latter enables learning lightweight adapted modules for downstream tasks.
Currently, a de facto paradigm for handling NLP tasks has been formed, dividing practitioners into two groups:
(1) upstream suppliers, who pre-train PLMs on task-agnostic data and release them on public platforms, e.g., HuggingFace~\citep{wolf-etal-2020-transformers}, and (2) downstream consumers, who download the PLM and conduct personalized adaptation using task-specific data. The corresponding adapted weights might then be shared with third parties via platforms such as AdapterHub~\citep{pfeiffer2020AdapterHub}.

\begin{figure}[!t]
\centering
\includegraphics[width=0.46\textwidth]{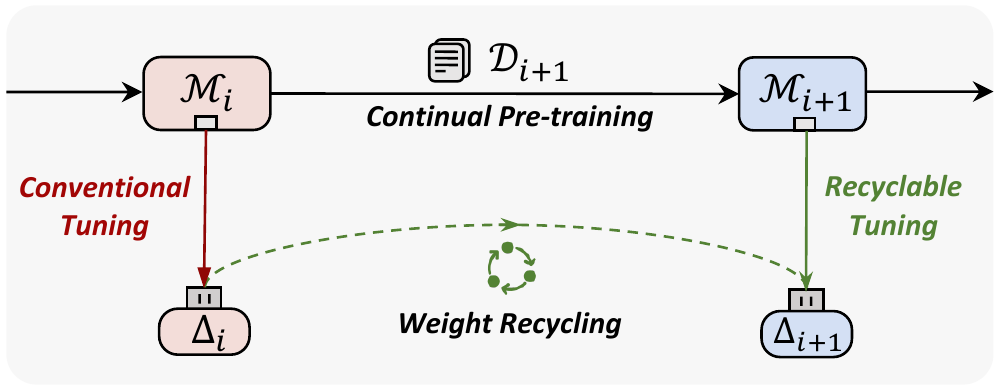}
\caption{Task formulation. The original PLM $\mathcal{M}_i$ is upgraded to $\mathcal{M}_{i+1}$ through continual pre-training on emerging data $\mathcal{D}_{i+1}$. Our goal is to recycle the existing adapted weights $\Delta_i$ of $\mathcal{M}_i$ for tuning $\mathcal{M}_{i+1}$.}
\label{fig:motivation}
\end{figure}

In real-world scenarios, PLMs may constantly get upgraded and released by the supplier. Correspondingly, the customer-side compatible update of adapted weights becomes necessary. \textit{Continual pre-training}~\citep{qin-etal-2022-elle} is a typical scenario where PLMs continually acquire fresh knowledge from growing data and gradually get upgraded. Before an upgraded PLM is released, consumers may have tuned the original PLM for various tasks and stored the adapted weights. However, when tuning the upgraded PLM, these outdated adapted weights will typically be ignored and discarded. This can lead to a loss of knowledge about downstream tasks encapsulated in the outdated weights, as well as a potential waste of computational resources. In this paper, we bring this issue to the forefront and argue that proper algorithms for recycling outdated adapted weights should be developed. To this end, we formulate the task of \textit{recyclable tuning for continual pre-training}, which is illustrated in Figure~\ref{fig:motivation}.

Due to the parameter change during continual pre-training, one potential concern for recycling outdated adapted weights is their mismatch with the upgraded PLM. However, our pilot studies reveal that directly applying the outdated weights to the upgraded PLM yields substantial performance improvements as compared to zero-shot inference of the PLM. This shows that the upgraded PLM remains compatible with the outdated weights to some extent, indicating a close connection between continually pre-trained PLMs. Intuitively, such a connection provides a strong basis for our assertion that outdated weights are recyclable and useful.

To uncover hints for solving our task, we further investigate such a connection from two aspects: (1) \textit{linear mode connectivity}~\citep{qin2022exploring}. We demonstrate that after adapting both the upgraded PLM and the original PLM to the same task, linearly interpolating the parameters of both adapted models could produce a series of checkpoints with high task performance (low loss). Such a property indicates a close parametric connection of both PLMs in the loss landscape; (2) \textit{functional similarity}. After adapting both PLMs to the same task, we observe that their corresponding attention heads exhibit similar patterns given the same input. Such representational proximity implies that both PLMs own similar functionalities during text processing. 

Both analyses above demonstrate the close connections between continually pre-trained PLMs. Based on the corresponding findings, we propose two methods for recyclable tuning:

(1) \textbf{Initialization-based method}, which leverages the adapted weights of the original PLM as the initialization for the upgraded PLM. This method is motivated by their close parametric connection in the loss landscape. We demonstrate that for a target task, initializing the tunable parameters with the outdated weights from a similar source task could accelerate the convergence and improve the training efficiency, compared to using random initialization. In addition, after sufficient training, this method generally improves the final performance. We also observe that the benefits of this method in terms of convergence and performance are greater when the source and target tasks are more similar.

(2) \textbf{Distillation-based method}, which distills the knowledge stored in outdated weights for tuning the upgraded PLM. We demonstrate that knowledge distillation can effectively facilitate knowledge transfer between continually pre-trained PLMs. Using only a small number of labeled examples, the upgraded PLM can outperform the original PLM when trained with far more examples. We also show that both initialization-based and distillation-based methods can be combined to further improve the performance. This means knowledge transfer through parameter space and model outputs are complementary to each other.

In a nutshell, these results highlight the practical benefits of recyclable tuning and point to an important future direction in sustainable NLP.
\section{Related Work}

\paragraph{Continual Pre-training.}
Conventionally, PLMs are trained on static data, ignoring that streaming data from various sources could continually grow. Continual pre-training requires PLMs to accumulate new knowledge in a continual manner~\citep{gururangan2020don}, meanwhile alleviating the catastrophic forgetting problem. Prior works in this field focus on building benchmarks and analyses~\citep{jang2021towards,jang2022temporalwiki}. Later works explored the applicability of traditional continual learning algorithms under this setting~\citep{jin-etal-2022-lifelong,wu2021pretrained}. Recent efforts were also spent on continual pre-training in a computationally efficient way~\citep{qin-etal-2022-elle}.

Previous works focus on improving the capabilities of PLMs during pre-training from the standpoint of upstream \textbf{suppliers}. Instead, we shift the focus to downstream adaptation from the perspective of \textbf{customers}. We highlight a previously overlooked issue of the incompatibility between upgraded PLMs and the existing adapted weights. For the first time, we examine the connections between continually pre-trained models and demonstrate the potential benefits of recycling outdated weights.

\paragraph{Knowledge Transfer for PLMs.}
Transfer learning for PLMs has gained increasing attention recently. Some works study task-level transferability for an \textbf{individual} PLM and find that fine-tuning on certain source tasks conduces to the performance on similar target tasks~\citep{vu-etal-2020-exploring,poth-etal-2021-pre,aghajanyan-etal-2021-muppet}. Differently, we also study cross-task knowledge transfer for \textbf{two different PLMs} under the continual pre-training scenario (\cref{sec:application_init}). Besides, researchers also investigate cross-model knowledge transfer. They try to recycle lightweight adapted weights of the same task between two \textbf{independently} pre-trained PLMs, e.g., PLMs with distinct data~\citep{su-etal-2022-transferability}. As we would show later, unlike independently trained PLMs, \textbf{continually} pre-trained PLMs are guaranteed close connections. This distinction determines our setting is unique to previous works and may require different solutions. 
\section{Problem Formulation}
\label{sec:task_formulation}
\paragraph{Continual Pre-training.}
Following \citet{qin-etal-2022-elle}, we simulate the scenario where \textbf{new data} from $4$ domains is gathered sequentially, i.e., biomedical papers (\textsc{Bio}, $\mathcal{D}_1$)~\citep{lo2019s2orc}, amazon reviews (\textsc{Rev}, $\mathcal{D}_2$)~\citep{he2016ups}, computer science papers (\textsc{CS}, $\mathcal{D}_3$)~\citep{lo2019s2orc}, and news articles (\textsc{Ns}, $\mathcal{D}_4$)~\citep{zellers2019defending}. Starting from the official $\text{RoBERTa}_\texttt{BASE}$~\citep{liu2019roberta} (denoted as $\mathcal{M}_0$), we continually pre-train $\mathcal{M}_0$ on $4$ domains. For each domain, we set the pre-training steps to $12.5$k and the batch size to $2048$. Denote $\mathcal{M}_i$ as the PLM that finishes training on $\mathcal{D}_i$, and $\mathcal{M}_i(t)$ as the PLM that starts from $\mathcal{M}_{i-1}$ and is trained on $\mathcal{D}_i$ for $t$ steps. We assume the suppliers only release the PLM that finishes training on each domain, i.e., $\{\mathcal{M}_\text{1}, \cdots, \mathcal{M}_\text{4}\}$ are developed and released. The pre-training details are described in \cref{sec:training_detail_lifelong_pretrain}.

\paragraph{Downstream Adaptation.}
At the same time, we have a set of downstream tasks to handle. To adapt $\mathcal{M}_i$ ($0 \!\le\! i \le 4$) towards a task $\mathcal{T}_j$, we conduct supervised training using the loss function $\mathcal{L}_{\mathcal{T}_j}$. Denote the pre-trained weights of $\mathcal{M}_{i}$ as $\theta_{i}^0$, we obtain its adapted weights $\Delta_{i}^{\mathcal{T}_j}$ for $\mathcal{T}_j$ after training. By assembling both $\theta_{i}^0$ and $\Delta_{i}^{\mathcal{T}_j}$, the resultant model $\theta_{i}^{\mathcal{T}_j} \!=\! \theta_{i}^0 \!\oplus\! \Delta_{i}^{\mathcal{T}_j}$ can be deployed to handle $\mathcal{T}_j$. Throughout this paper, we consider two tuning methods: full-parameter fine-tuning and a representative parameter-efficient tuning method, adapter tuning~\citep{pmlr-v97-houlsby19a} (see \cref{sec:pet} for more backgrounds). For the former, we have $|\Delta_{i}^{\mathcal{T}_j}| \!=\! |\theta_{i}^0|$; while for the latter, $|\Delta_{i}^{\mathcal{T}_j}| \!\ll\! |\theta_{i}^0|$, where $|\cdot|$ denotes the number of parameters.

\paragraph{Recyclable Tuning.}
Before the release of an upgraded PLM $\mathcal{M}_{i'}$ ($i \!<\! i'$), we have obtained adapted weights $\Delta_{i}^{\mathcal{T}_j}$ of an old PLM $\mathcal{M}_{i}$ for task $\mathcal{T}_j$.
Recyclable tuning aims at transferring the knowledge of $\Delta_{i}^{\mathcal{T}_j}$ to assist tuning $\mathcal{M}_{i'}$ (i.e., learning new weights $\Delta_{i'}^{\mathcal{T}_j}$). We denote the above process as $\Delta_{i}^{\mathcal{T}_j} \!\rightarrow\! \Delta_{i'}^{\mathcal{T}_j}$. Intuitively, $\Delta_{i}^{\mathcal{T}_j}$ encapsulates abundant knowledge about the task $\mathcal{T}_j$, which should benefit learning $\Delta_{i'}^{\mathcal{T}_j}$ if exploited properly. Such benefits may include improving training efficiency or performance. To gain insights of solving the task, we first conduct a series of empirical analyses in \cref{sec:empirical_study} to understand the connections among $\mathcal{M}_{i}$, $\mathcal{M}_{i'}$, $\Delta_{i}^{\mathcal{T}_j}$, and $\Delta_{i'}^{\mathcal{T}_j}$.

\section{Empirical Analysis}
\label{sec:empirical_study}
We first investigate the compatibility of outdated weights and the upgraded PLM (\cref{sec:direct_apply}), then we explore the (1) parametric connections and (2) representational connections of continually pre-trained PLMs from two aspects: (1) linear mode connectivity (\cref{sec:parametric_connection}) and (2) functional similarity (\cref{sec:functional_analysis}). The implementation details are left in \cref{sec:training_detail_empirical}.

\subsection{Model Compatibility Analysis}
\label{sec:direct_apply}
We explore to what extent the outdated weights are compatible with the upgraded PLM and how this compatibility changes during continual pre-training. Specifically, we directly apply outdated weights to the upgraded PLM and record the performance variation during continual pre-training.

\paragraph{Settings.}
We first investigate the process when upgrading $\mathcal{M}_0$ to $\mathcal{M}_1$ on the \textsc{Bio} domain ($\mathcal{D}_1$). For downstream evaluation, we choose two classification tasks: \textsc{ChemProt}~\citep{kringelum2016chemprot}, which is a relevant downstream task to the \textsc{Bio} domain, and $\textsc{MNLI}$~\citep{williams-etal-2018-broad}. Denote the model continually pre-trained on $\mathcal{D}_1$ for $t$ steps as $\mathcal{M}_1(t)$, its pre-trained weights as $\theta_{1}^0(t)$, and the adapted weights of $\mathcal{M}_0$ for the downstream task as $\Delta_{0}^\mathcal{T}$. We directly apply $\Delta_{0}^\mathcal{T}$ to the upgraded PLM $\mathcal{M}_1(t)$, i.e., $\theta_{1}^0(t) \oplus \Delta_{0}^\mathcal{T}$, and evaluate the performance on the test set of the downstream task. In experiments, $t$ is selected from $1.25$k to $12.5$k with an interval of $1.25$k. We also report $\mathcal{M}_1(t)$'s zero-shot inference performance by testing $\theta_{1}^0(t)$.

\begin{figure}[!t]
    \centering
    \subfigure{\includegraphics[width=0.48\textwidth]{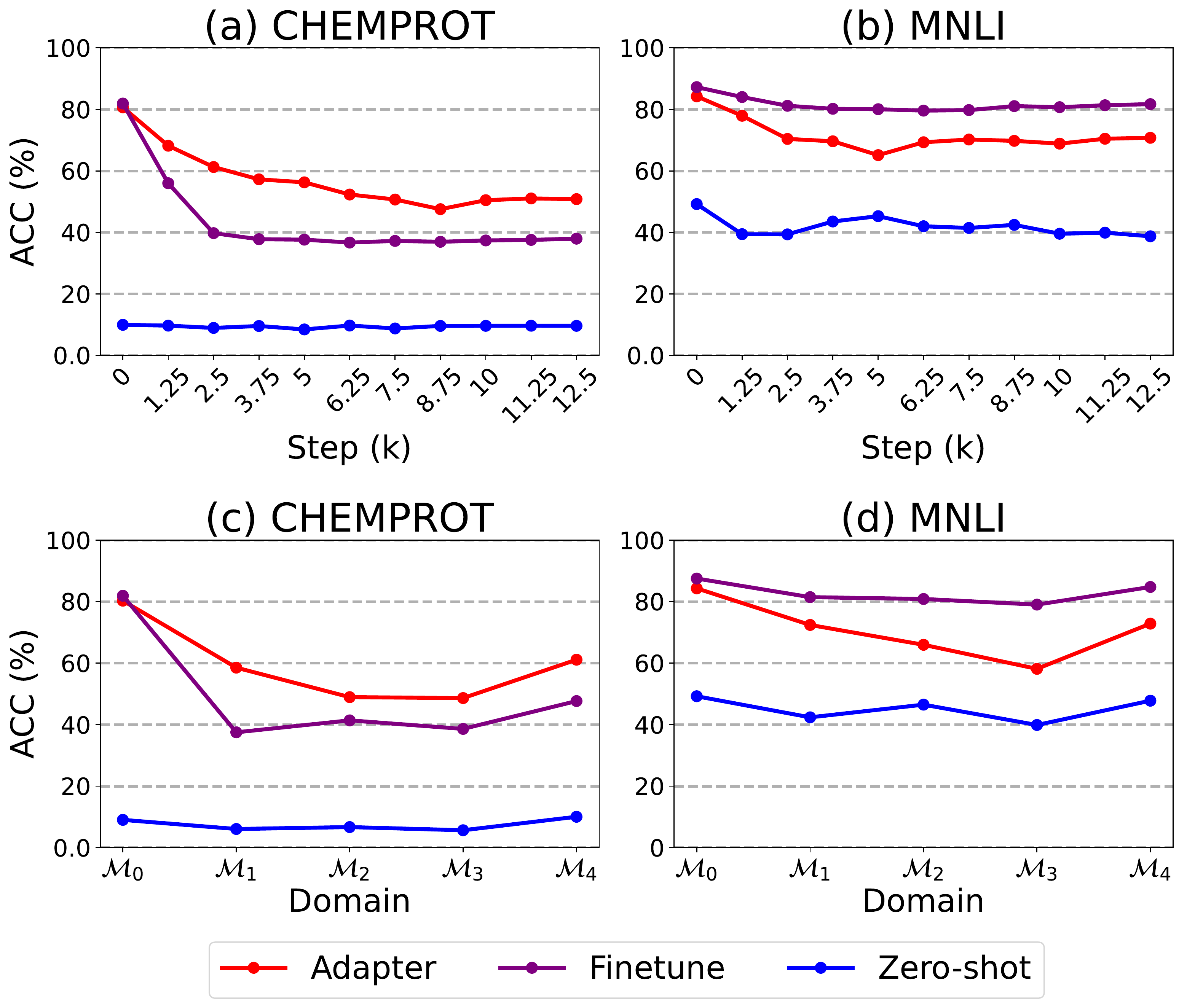}} 
    \caption{(a, b): performance variation w.r.t. pre-training steps ($t$) when applying the outdated weights ($\Delta_{0}^\mathcal{T}$) to $\mathcal{M}_1(t)$. (c, d): performance variation when applying the outdated weights ($\Delta_{0}^\mathcal{T}$) to $\{\mathcal{M}_1, \cdots, \mathcal{M}_4\}$.}
    \label{fig:direct_apply}
\end{figure}

\paragraph{Results.}
From the results in Figure~\ref{fig:direct_apply} (a, b), we observe that for both adapter and fine-tuning: (1) with $t$ increasing, the performance of $\theta_{1}^0(t) \!\oplus\! \Delta_{0}^\mathcal{T}$ drops quickly at first. This means that $\Delta_{0}^\mathcal{T}$ becomes outdated shortly after the backbone model $\mathcal{M}_1(t)$ changes. (2) After sufficient pre-training steps, the performance converges to a plateau which is still much higher than the zero-shot inference performance of $\mathcal{M}_1(t)$. This implies that \textbf{continually pre-trained PLMs are intrinsically connected with their ``ancestors''}, otherwise the ancestor's adapted weights $\Delta_0^\mathcal{T}$ would not improve the performance of its offspring $\mathcal{M}_1(t)$.

\paragraph{Extension to Multiple Domains.}
Next, we extend the above experiments to $4$ sequentially released PLMs as mentioned in \cref{sec:task_formulation} by directly applying $\Delta_{0}^\mathcal{T}$ to $\{\mathcal{M}_1, \cdots, \mathcal{M}_4\}$. We derive from Figure~\ref{fig:direct_apply} (c, d) that: (1) applying outdated weights consistently performs better than zero-shot inference even if the backbone PLM is trained over multiple domains; (2) the performance of $\mathcal{M}_4$ is the best among $\{\mathcal{M}_1, \cdots, \mathcal{M}_4\}$ though $\mathcal{M}_4$ is trained for the longest time. This may be because the \textsc{Ns} domain ($\mathcal{D}_4$) is the most similar one to $\mathcal{M}_0$'s pre-training data~\citep{gururangan2020don}, and \textbf{continual pre-training on a similar domain of the original PLM mitigates the incompatibility}.

\subsection{Linear Mode Connectivity Analysis}
\label{sec:parametric_connection}
\paragraph{Backgrounds.}
Linear mode connectivity measures whether two sets of model weights can be connected via a linear parametric path, along which the performance (loss) of the downstream task remains high (low)~\citep{frankle2020linear}. In other words, it tests whether linear interpolations of two model weights perform comparably to both endpoints. If this property holds, then both model weights probably lie in the same loss basin, which indicates a close connection between them in the parameter space~\citep{qin2022exploring}. For more detailed backgrounds, please refer to \cref{sec:mode_connectivity}.

\paragraph{Settings.} Following most of the settings in \cref{sec:direct_apply}, we adapt both $\mathcal{M}_{0}$ and $\mathcal{M}_1(t)$ towards the task \textsc{ChemProt} and obtain the weights $\theta_{0}^\mathcal{T}$ and $\theta_{1}^\mathcal{T}(t)$, where $\theta_{0}^\mathcal{T} = \theta_{0}^0 \oplus \Delta_{0}^\mathcal{T}$ and $\theta_{1}^\mathcal{T}(t) = \theta_{1}^0(t) \oplus \Delta_{1}^\mathcal{T}(t)$. Then we linearly interpolate both $\theta_{0}^\mathcal{T}$ and $\theta_{1}^\mathcal{T}(t)$ as:
\begin{equation}
\label{eq:linear}
\begin{aligned}
    \theta(\mu) = (1-\mu) \theta_{0}^\mathcal{T} + \mu \theta_{1}^\mathcal{T}(t), 
\end{aligned}
\end{equation}
where $\mu \in (0, 1)$. In experiments, we evaluate the performance of $25$ evenly distributed interpolations and two endpoints (i.e., $\mu = 0$ and $\mu = 1$). If there does not exist a significant performance drop along the linear path, we deem both endpoints linearly mode connected. We choose $\mathcal{M}_1(t)$ that is continually pre-trained for $\{2.5, 5.0, 7.5, 10.0, 12.5\}$k steps and evaluate mode connectivity for each $\mathcal{M}_1(t)$ and $\mathcal{M}_0$. In addition, we pre-train a new $\text{RoBERTa}_{\texttt{BASE}}$ (dubbed as $\mathcal{M}_\text{IND}$) from scratch (details in \cref{sec:training_detail_lifelong_pretrain}) and test its connectivity with $\mathcal{M}_{0}$, i.e., $\theta(\mu) \!=\! (1\!-\!\mu) \theta_{0}^\mathcal{T} \!+\! \mu \theta_{\text{IND}}^\mathcal{T}$. In this way, we can compare the difference between continually pre-trained models ($\mathcal{M}_{0}$ and $\mathcal{M}_1(t)$) and independently pre-trained models ($\mathcal{M}_{0}$ and $\mathcal{M}_\text{IND}$).

\begin{figure}[!t]
    \centering
    \subfigure{\includegraphics[width=0.48\textwidth]{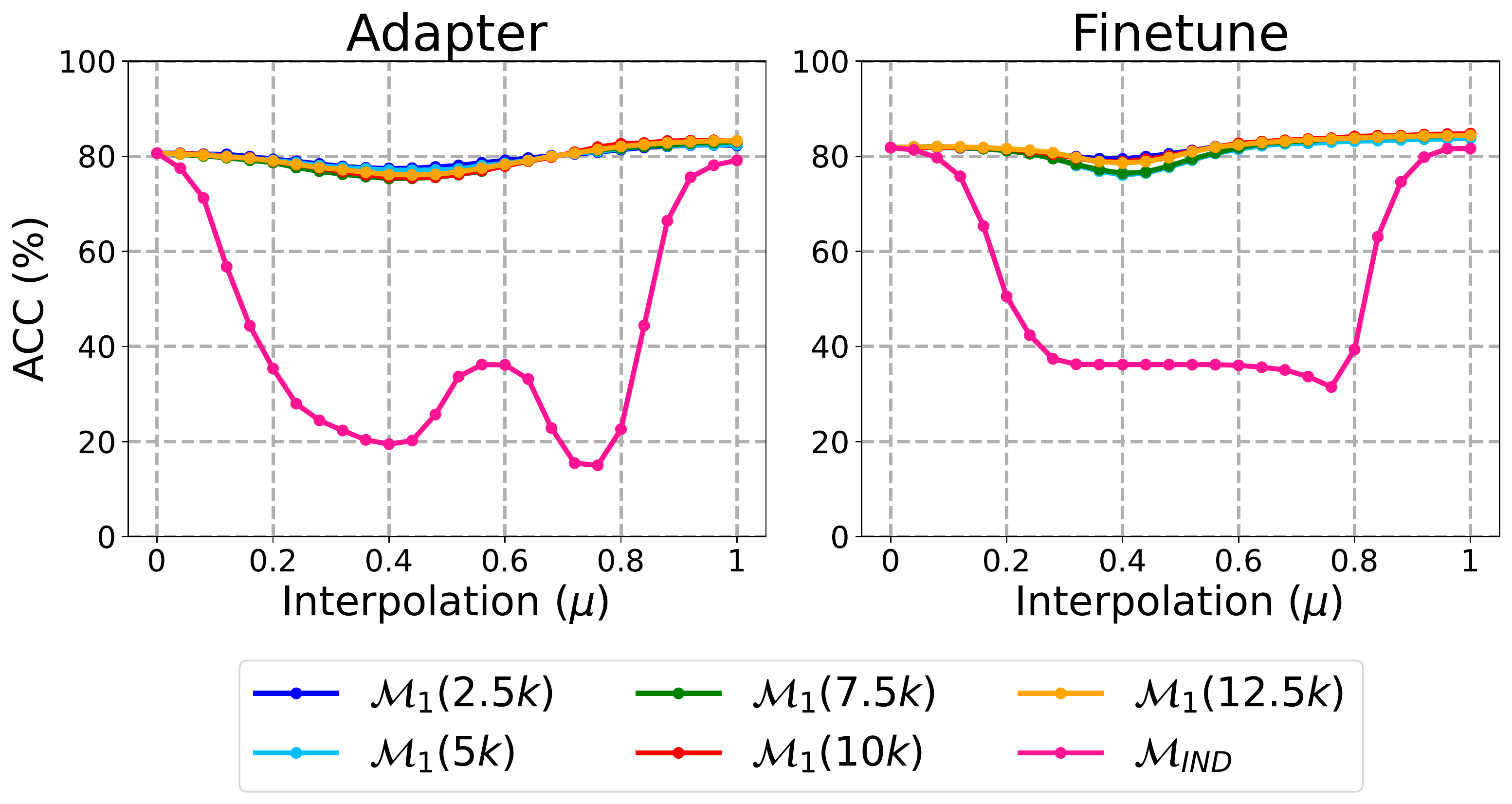}} 
    \caption{The performance of linear interpolations between two adapted PLMs on \textsc{ChemProt}. $\mu=0$ means $\mathcal{M}_{0}$, and $\mu=1$ means $\mathcal{M}_{1}(t)$ or $\mathcal{M}_\text{IND}$.}
    \label{fig:mode_connectivity}
\end{figure}

\paragraph{Results.}
We illustrate the performance of the interpolations and two endpoints in Figure~\ref{fig:mode_connectivity}, from which we conclude that: (1) for continually pre-trained PLMs, although there exists a small performance drop in the midpoint, the interpolations generally achieve comparable performance to endpoints; (2) the connectivity does not vary much with $t$ increasing, which means within a reasonable range, the connectivity is not sensitive to longer pre-training; (3) while for independently trained PLMs, the performance drops significantly in the middle, which means the adapted weights of these PLMs cannot be linked by a high-performance linear path; (4) the above conclusions hold for both adapter and fine-tuning.

The above findings imply that when learning the same task, \textbf{two continually pre-trained PLMs would probably be optimized into two minima lying in the same loss basin}, or at least the optimal regions corresponding to both minima have a substantial intersection; otherwise, there should exist a significant performance drop in between.

\begin{figure}[!t]
    \centering
    \subfigure{\includegraphics[width=0.48\textwidth]{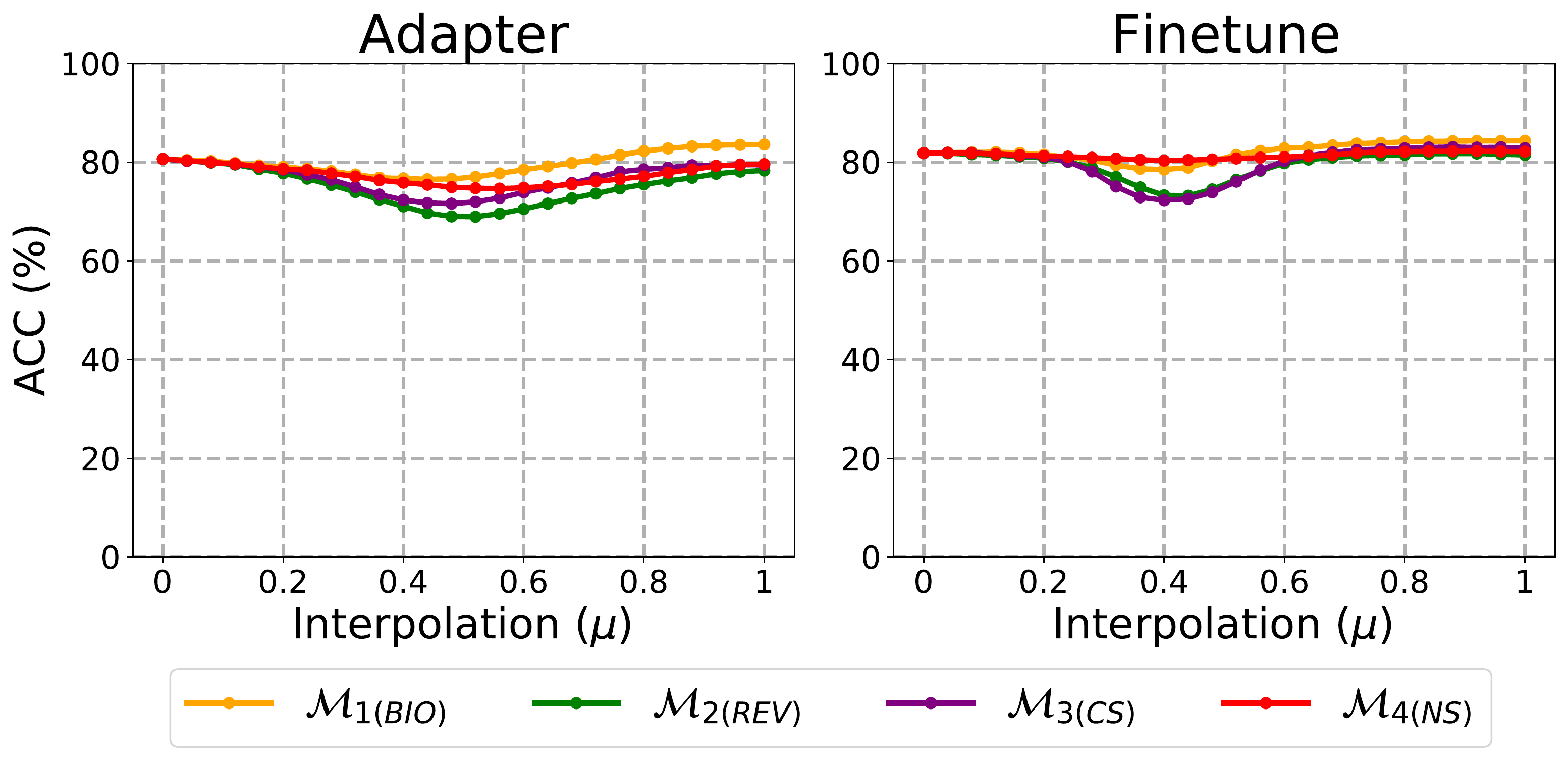}} 
    \caption{Linear mode connectivity between the initial $\mathcal{M}_0$ ($\mu=0$) and $4$ sequentially pre-trained PLMs $\{\mathcal{M}_1, \cdots, \mathcal{M}_4\}$ over multiple domains ($\mu=1$).}
    \label{fig:mode_connectivity_multiple_domain}
\end{figure}

Intuitively, the existence of a high-performance (low-loss) path between two optimal regions implies that \textbf{model weights can be easily optimized from one optimal region to another without incurring a loss barrier}. In this regard, it is promising to use outdated adapted weights as the initialization to find the optimal solution for the upgraded PLM, which would be explored in \cref{sec:application_init}. In this way, we explicitly facilitate cross-model knowledge transfer through the parameter space.

\paragraph{Extension to Multiple Domains.}
Next, we evaluate linear mode connectivity between the initial $\mathcal{M}_0$ and $\mathcal{M}_i$ ($1 \!\le\! i \!\le\! 4$) using the task \textsc{ChemProt}. We derive from the results in Figure~\ref{fig:mode_connectivity_multiple_domain} that although the performance tends to drop slightly near the midpoint, the connectivity of all continually pre-trained models is still far better than independent PLMs (i.e., $\mathcal{M}_\text{IND}$ in Figure~\ref{fig:mode_connectivity}). We also observe that the performance drop between $\mathcal{M}_0$ and $\mathcal{M}_2$ is larger than $\mathcal{M}_0$ and $\mathcal{M}_4$, though $\mathcal{M}_4$ is trained for a longer time than $\mathcal{M}_2$. This means \textbf{longer pre-training does not necessarily result in poorer connectivity; rather, the pre-training domain has a great impact}. 

\subsection{Functional Similarity Analysis}
\label{sec:functional_analysis}
The close parametric connection revealed by linear mode connectivity does not guarantee that continually pre-trained PLMs share similar functionalities when processing the text information. Following \citet{pmlr-v97-gong19a}, we explore functional similarity through the lens of attention distribution.
Specifically, we investigate three continually pre-trained models ($\mathcal{M}_0$, $\mathcal{M}_1$, and $\mathcal{M}_2$) and fine-tune them on \textsc{ChemProt} to obtain adapted models ($\theta_0^{\mathcal{T}}$, $\theta_1^{\mathcal{T}}$, and $\theta_2^{\mathcal{T}}$). We feed the same input sampled from \textsc{ChemProt} to the three adapted models. Then we select attention heads from the same position (i.e., the $h$-th head in the $l$-th layer) in three models, and visualize their attention distribution. Note the selected head of $\mathcal{M}_{i+1}$ is trained from that of $\mathcal{M}_i$.

\begin{figure}[!t]
    \centering
    \subfigure{\includegraphics[width=0.45\textwidth]{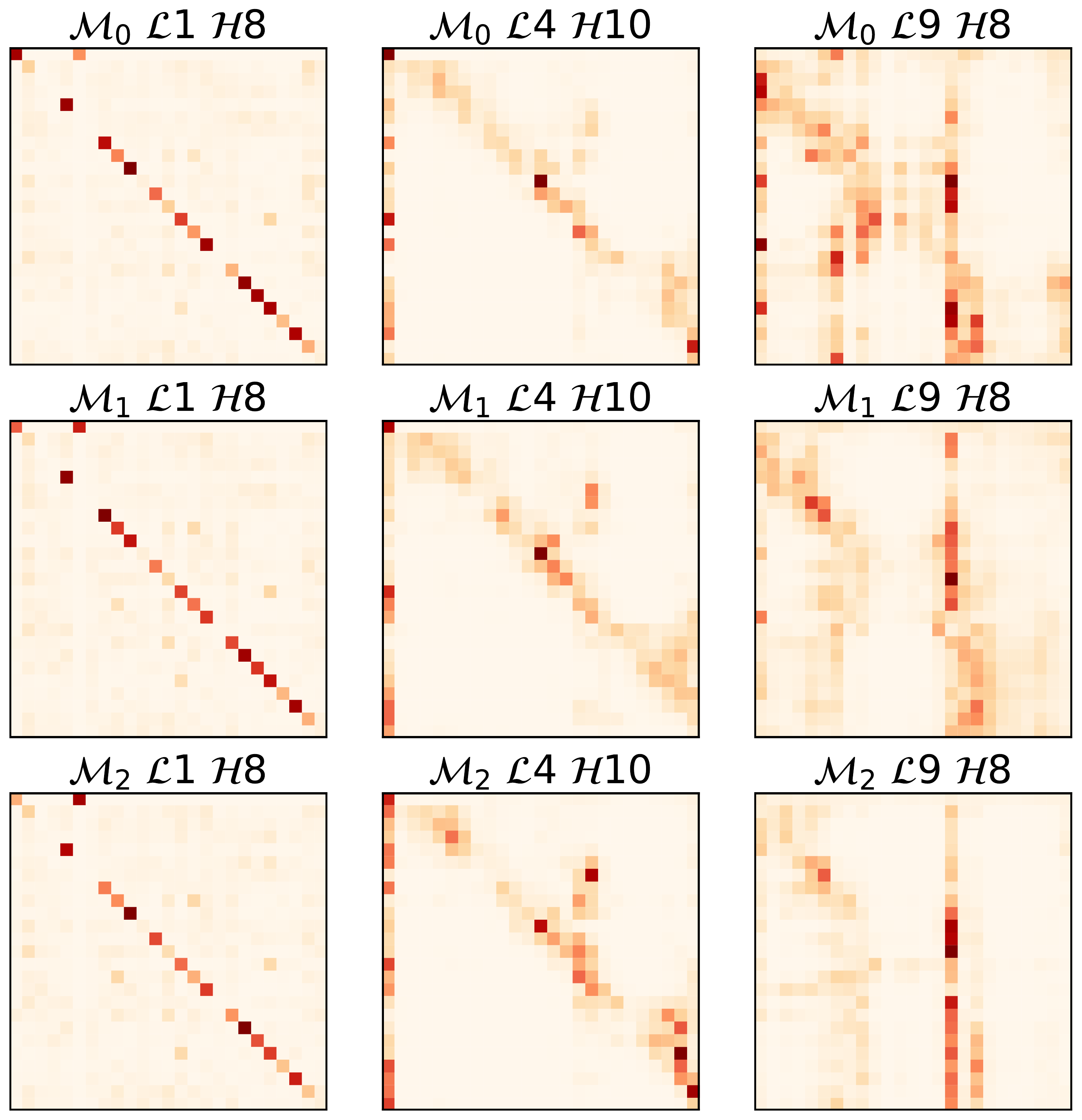}}
    \caption{Visualization of attention heads in fine-tuned $\mathcal{M}_0$, $\mathcal{M}_1$, and $\mathcal{M}_2$ given the same input. For instance, ``$\mathcal{L}4$ $\mathcal{H}10$'' refers to the $10$-th head in the $4$-th layer. An attention head of $\mathcal{M}_{i+1}$ is trained from that of $\mathcal{M}_i$ in the same column. In the heatmap, the color of the $i$-th element in the $j$-th row indicates the attention value from the $j$-th token to the $i$-th token. For more visualizations (including $\mathcal{M}_\text{IND}$), please refer to \cref{sec:more_visualziation_attention}.}
    \label{fig:attention}
\end{figure}

From Figure~\ref{fig:attention}, it is found that the attention patterns of $\mathcal{M}_1$ and $\mathcal{M}_2$ are quite similar to those of their ``ancestor'' $\mathcal{M}_0$. Such representational proximity indicates that \textbf{the corresponding modules of continually pre-trained PLMs own similar functionalities}. Since adapted weights play a pivotal role in stimulating PLM's abilities and functionalities~\citep{ding2022delta}, such functional similarity partially explains why the outdated adapted weights can be directly applied to the upgraded PLM and achieve non-trivial performance in \cref{sec:direct_apply}.

In a nutshell, all the analyses in this section validate the close connection between continually pre-trained PLMs. Intuitively, such a connection implies that the adaptation process of these PLMs towards downstream tasks should be closely related and transferable as well, which serves as the strong basis for our recyclable tuning.
\section{Methods and Experiments}
\label{sec:application}
Based on the findings in \cref{sec:empirical_study}, we propose two ways to explore the practical benefits of recyclable tuning: initialization-based method (\cref{sec:application_init}) and distillation-based method (\cref{sec:application_distil}). The training details of this section are discussed in \cref{sec:training_detail_application}.

\subsection{Initialization-based Recyclable Tuning}
\label{sec:application_init}

We first investigate directly using outdated weights as the initialization for tuning the upgraded PLM.

\begin{figure}[!t]
    \centering
    \subfigure{\includegraphics[width=0.45\textwidth]{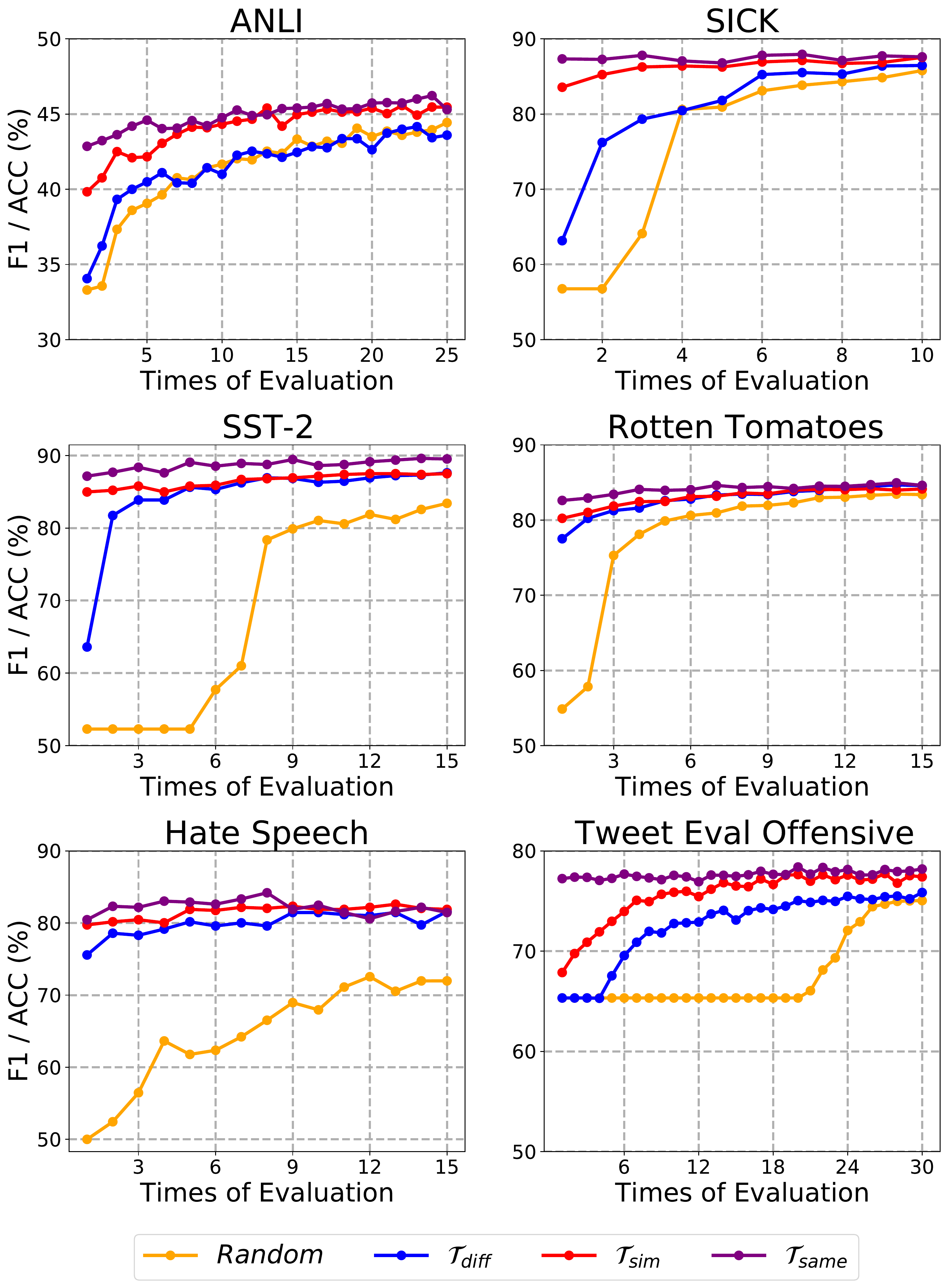}} 
    \caption{The performance variation in the early stage of adapter tuning for $6$ target tasks from different initialization. Different tasks are evaluated with different intervals of training steps, see \cref{sec:training_detail_application} for details.}
    \label{fig:init}
\end{figure}

\paragraph{Framework.} Without loss of generality, we experiment when the initial PLM $\mathcal{M}_0$ is continually pre-trained on the \textsc{Bio} domain ($\mathcal{D}_1$) and upgraded to $\mathcal{M}_1$. Before the release of a new PLM $\mathcal{M}_1$, assume we have tuned $\mathcal{M}_0$ on N tasks $\{\mathcal{T}_0, \cdots, \mathcal{T}_\text{N}\}$ and obtained the corresponding adapted weights $\{\Delta_0^{\mathcal{T}_1}, \cdots, \Delta_0^{\mathcal{T}_\text{N}}\}$. When tuning $\mathcal{M}_1$ on a target task $\mathcal{T}_t$, instead of using the random initialization for tunable weights, we initialize them using $\mathcal{M}_0$'s adapted weights $\Delta_{0}^{\mathcal{T}_s}$ trained on a source task $\mathcal{T}_s$.

Considering that in practice, it is possible that the outdated weights of exactly the same task are not available, i.e., $\mathcal{T}_t \!\neq\! \mathcal{T}_s$. Thus we explore whether initialization from the outdated weights of a different task would suffice for our goal. Specifically, we consider three types of source tasks: (1) $\mathcal{T}_{\text{same}}$, which is the \textit{same} task as the target one; (2) $\mathcal{T}_{\text{sim}}$, which denotes a task \textit{similar} to $\mathcal{T}_t$, both $\mathcal{T}_{\text{sim}}$ and $\mathcal{T}_t$ typically belong to the same task type; (3) $\mathcal{T}_{\text{diff}}$, which belongs to a \textit{different} task category from $\mathcal{T}_t$.

\paragraph{Settings.}
We experiment with $6$ \textbf{target tasks} of $3$ types: (1) \textit{natural language inference}: ANLI~\citep{nie-etal-2020-adversarial} and SICK~\citep{marelli2014sick}, (2) \textit{sentiment analysis}: SST-2~\citep{socher-etal-2013-recursive} and Rotten Tomatoes~\citep{pang-lee-2005-seeing}, (3) \textit{emotion detection}: Hate Speech~\citep{hateoffensive} and Tweet Eval-Offensive~\citep{barbieri-etal-2020-tweeteval}. The choices of $\mathcal{T}_{\text{sim}}$ and $\mathcal{T}_{\text{diff}}$ for each target task are listed in Table~\ref{tab:convergence_task} in the appendix. 

We compare the proposed initialization strategies with random initialization and record (1) the test performance variation (w.r.t. training steps) during the early stage of downstream adaptation (Figure~\ref{fig:init}), and (2) the best test performance after the adaptation converges (Table~\ref{tab:init}). For adaptation, we mainly investigate adapter tuning and leave the experiments of fine-tuning in \cref{sec:additional_exp_initialization}.

\paragraph{Results.}
The observations and corresponding conclusions are summarized as follows:

\begin{table}[!t]
  \centering
  \small
    \begin{tabular}{c@{~~}|c@{~~}c@{~~}c@{~~}c@{~~}}
    \toprule
    \textbf{Initialization} & \textit{Random} & $\mathcal{T}_{\text{diff}}$ & $\mathcal{T}_{\text{sim}}$ & $\mathcal{T}_{\text{same}}$ \\
    \midrule
    ANLI  & $42.5_{\pm0.7}$  & $43.7_{\pm0.7}$  &  $46.2_{\pm0.8}$    & $\textbf{46.8}_{\pm0.1}$  \\
    SICK  & $87.6_{\pm0.6}$  & $\textbf{88.1}_{\pm0.7}$  &  $87.7_{\pm0.2}$  & $87.9_{\pm0.6}$  \\
    SST-2 & $89.3_{\pm0.5}$  & $89.8_{\pm0.5}$  &  $89.1_{\pm0.7}$  & $\textbf{90.3}_{\pm0.5}$  \\
    R. Tomatoes & $85.3_{\pm0.2}$  & $85.2_{\pm0.2}$  &  $\textbf{85.5}_{\pm0.2}$  & $84.7_{\pm0.4}$  \\
    H. Speech & $77.2_{\pm0.6}$  &   $80.1_{\pm0.8}$    &  $81.9_{\pm1.3}$   & $\textbf{83.9}_{\pm0.5}$  \\
    T. Offensive & $84.2_{\pm0.6}$  &  $84.0_{\pm0.4}$   &  $\textbf{84.8}_{\pm0.2}$   & $83.8_{\pm0.1}$  \\
    \midrule
    Avg. & $77.7_{\pm0.5}$  &   $78.5_{\pm0.6}$   &   $79.2_{\pm0.6}$    &  $\textbf{79.6}_{\pm0.4}$ \\
    \bottomrule
    \end{tabular}%
  \caption{The best test performance on $6$ target tasks with adapter tuning from different initialization.}
  \label{tab:init}%
\end{table}

(1) \textbf{Faster convergence}: we observe from Figure~\ref{fig:init} that compared with the random initialization baseline, our method significantly accelerates the convergence of downstream adaptation. This suggests that the outdated weights provide a more effective initialization, allowing the PLM to be more easily optimized to the desired local optima. In practice, this method could improve the training efficiency of tuning the upgraded PLM, which saves the computations needed for adaptation.

(2) \textbf{Improved task performance}: we also conclude from Table~\ref{tab:init} that after sufficient training, initialization from the outdated weights of each type of source tasks (even for $\mathcal{T}_\text{diff}$) could improve the final performance (up to $+1.9$ average improvement). This demonstrates that initialization serves as a valid way for cross-model knowledge transfer.

(3) \textbf{Similar source tasks benefit more}: comparing the results of initialization from different source tasks, we find that the improvement in both convergence and performance can be generally ranked as $\mathcal{T}_\text{same} \!>\! \mathcal{T}_\text{sim} \!>\! \mathcal{T}_\text{diff}$. This is because the knowledge required by more similar tasks has a greater overlap. Thus the knowledge transfer benefits more when the target task and source task are more similar. In practice, this finding expands the selection scope of source adapted weights, broadening the application scenarios for our initialization-based method.

\subsection{Distillation-based Recyclable Tuning}
\label{sec:application_distil}
According to \citet{lin-etal-2021-weight}, model outputs often contain sufficient supervision that is complementary to the knowledge stored in parameters. Therefore, besides the initialization-based method, we also explore knowledge distillation~\citep{hinton2015distilling} to recycle the outdated weights. 
\paragraph{Framework.} Given a task $\mathcal{T}_j$, assume we have optimized an outdated PLM $\mathcal{M}_i$ and obtained its adapted weights $\Delta_i^{\mathcal{T}_j}$. Our goal is to distill the knowledge stored in $\Delta_i^{\mathcal{T}_j}$ to optimize an updated PLM $\mathcal{M}_{i+1}$. We follow \citet{sun-etal-2019-patient} to construct our framework. For each data point $x$ from $\mathcal{T}_j$, denote $\mathcal{P}(x, \theta_i^{\mathcal{T}_j})$ as the probability distribution the adapted $\mathcal{M}_i$ assigns over the label space, where $\theta_i^{\mathcal{T}_j} \!=\! \theta_i^0 \!\oplus\! \Delta_i^{\mathcal{T}_j}$. We minimize the KL divergence between probabilities predicted by $\mathcal{M}_i$ and $\mathcal{M}_{i+1}$. In addition, $\mathcal{M}_{i+1}$ mimics $\mathcal{M}_i$'s intermediate hidden representations of each layer. Specifically, given the same input $x$, denote $\mathbf{h}_k(x, \theta_i^{\mathcal{T}_j})$ and $\mathbf{h}_k(x, \theta_{i+1}^{\mathcal{T}_j})$ as the normalized hidden states of the $k$-th layer of $\mathcal{M}_{i}$ and $\mathcal{M}_{i+1}$, we minimize the mean-square loss of hidden states together with the KL divergence as follows:
\begin{equation}
\begin{aligned}
    &\mathcal{L}_{\text{KD}} = \mathrm{KL}(\mathcal{P}(x, \theta_i^{\mathcal{T}_j}) || \mathcal{P}(x, \theta_{i+1}^{\mathcal{T}_j})) + \\
    & \alpha \Sigma_{k} ||\mathbf{h}_k(x, \theta_i^{\mathcal{T}_j}) \!-\! \mathbf{h}_k(x, \theta_{i+1}^{\mathcal{T}_j})||^2,
\end{aligned}
\end{equation}
where $\alpha$ denotes a hyper-parameter. \textbf{During optimization, only $\Delta_{i+1}^{\mathcal{T}_j}$ is tunable}. Besides $\mathcal{L}_\text{KD}$, we also introduce the original task loss $\mathcal{L}_{\mathcal{T}_j}$, which is calculated using supervised training examples from task $\mathcal{T}_j$, with another hyper-parameter $\beta$:
\begin{equation}
\begin{aligned}
    \mathcal{L}_{\text{final}} = \beta \mathcal{L}_{\mathcal{T}_j} + (1 - \beta) \mathcal{L}_{\text{KD}}.
\end{aligned}
\label{eq:distillation-based}
\end{equation}

\paragraph{Settings.}
We consider the sequentially released PLMs $\{\mathcal{M}_0, \cdots, \mathcal{M}_4\}$ as mentioned in \cref{sec:task_formulation}. Following \citet{gururangan2020don}, we choose three tasks $\mathcal{T}_1$: \textsc{ChemProt}, $\mathcal{T}_2$: \textsc{IMDB}~\citep{maas-etal-2011-learning} and $\mathcal{T}_3$: \textsc{ACL-ARC}~\citep{jurgens2018measuring}, which are relevant to domain $\mathcal{D}_1$, $\mathcal{D}_2$ and $\mathcal{D}_3$, respectively. We mainly consider recyclable tuning between adjacent PLMs, i.e., $\mathcal{M}_i$ and $\mathcal{M}_{i+1}$, and also evaluate non-adjacent PLMs (e.g., $\mathcal{M}_i$ and $\mathcal{M}_{i+2}$) in \cref{sec:non_adjacent}. For each task $\mathcal{T}_i$ ($i \in \{1,2,3\}$), we consider two settings:

(a) First, we recycle $\mathcal{M}_{i}$'s outdated weights to $\mathcal{M}_{i+1}$, which is denoted as $\Delta_i^{\mathcal{T}_i} \!\rightarrow\! \Delta_{i+1}^{\mathcal{T}_i}$. Here the evaluated task $\mathcal{T}_i$ is relevant to the pre-training domain $\mathcal{D}_i$ of the original PLM $\mathcal{M}_i$. During continual pre-training on $\mathcal{D}_{i+1}$, $\mathcal{M}_{i+1}$ suffers from catastrophic forgetting of $\mathcal{D}_i$. Hence $\mathcal{M}_{i+1}$ should perform worse on $\mathcal{T}_i$ than $\mathcal{M}_{i}$. In experiments, both PLMs are adapted using the same $32$-shot dataset.

\begin{table}[!tbp]
  \centering
  \small
    \begin{tabular}{c@{~}c@{~~}c@{~}c@{~}c@{~}c@{~}}
    \toprule
    \multicolumn{2}{c}{\textbf{Method}} & $\textit{Teacher}$ & $\mathcal{L}_\text{final}$-$\mathcal{L}_\text{KD}$  & $\mathcal{L}_\text{final}$ & $\mathcal{L}_\text{final}$+\textit{Init.} \\
    \midrule
    \multicolumn{6}{c}{\textit{Setting (a):} $\Delta_{i}^{\mathcal{T}_i} \!\rightarrow\! \Delta_{i+1}^{\mathcal{T}_i}$, $i \in \{1,2,3\}$} \\
    \midrule
    \multirow{2}[2]{*}{$\Delta_{1}^{\mathcal{T}_1} \!\rightarrow\! \Delta_{2}^{\mathcal{T}_1}$} & \textbf{AP}    & $\textbf{65.2}_{\pm1.7}$ &   $58.0_{\pm0.9}$    & $62.4_{\pm1.3}$ & $63.8_{\pm3.2}$ \\
          & \textbf{FT}    & $\textbf{66.0}_{\pm1.4}$ &   $61.4_{\pm3.1}$    & $64.5_{\pm0.5}$ & $64.7_{\pm0.6}$ \\
    \midrule
    \multirow{2}[2]{*}{$\Delta_{2}^{\mathcal{T}_2} \!\rightarrow\! \Delta_{3}^{\mathcal{T}_2}$} & \textbf{AP}    & $\textbf{84.8}_{\pm1.3}$ &   $78.3_{\pm1.4}$    & $80.7_{\pm0.3}$ & $80.8_{\pm0.7}$ \\
          & \textbf{FT}    & $\textbf{82.0}_{\pm1.8}$ &  $76.7_{\pm2.2}$     & $79.5_{\pm1.5}$ & $79.7_{\pm1.9}$ \\
    \midrule
    \multirow{2}[2]{*}{$\Delta_{3}^{\mathcal{T}_3} \!\rightarrow\! \Delta_{4}^{\mathcal{T}_3}$} & \textbf{AP}    & $50.6_{\pm3.0}$ &  $48.2_{\pm2.9}$     & $48.0_{\pm1.4}$ & $\textbf{55.9}_{\pm3.9}$ \\
          & \textbf{FT}    & $52.5_{\pm0.6}$ &   $51.8_{\pm4.2}$    & $54.2_{\pm0.7}$ & $\textbf{61.3}_{\pm2.9}$ \\
    \midrule
    \multicolumn{6}{c}{\textit{Setting (b):} $\Delta_{i-1}^{\mathcal{T}_i} \!\rightarrow\! \Delta_{i}^{\mathcal{T}_i}$, $i \in \{1,2,3\}$} \\
    \midrule
    \multirow{2}[2]{*}{$\Delta_{0}^{\mathcal{T}_1} \!\rightarrow\! \Delta_{1}^{\mathcal{T}_1}$} & \textbf{AP}    & $59.1_{\pm2.5}$ &   $53.1_{\pm0.7}$    & $61.4_{\pm1.1}$ & $\textbf{64.7}_{\pm0.4}$ \\
          & \textbf{FT}    & $61.8_{\pm1.3}$ &   $56.6_{\pm1.2}$    & $59.3_{\pm1.5}$ & $\textbf{63.4}_{\pm0.7}$ \\
    \midrule
    \multirow{2}[2]{*}{$\Delta_{1}^{\mathcal{T}_2} \!\rightarrow\! \Delta_{2}^{\mathcal{T}_2}$} & \textbf{AP}    & $83.1_{\pm0.3}$ &   $84.8_{\pm1.3}$    & $86.0_{\pm0.2}$ & $\textbf{87.3}_{\pm0.4}$ \\
          & \textbf{FT}    & $83.3_{\pm0.6}$ &  $82.0_{\pm1.8}$     & $85.5_{\pm0.8}$ & $\textbf{86.8}_{\pm0.7}$ \\
    \midrule
    \multirow{2}[2]{*}{$\Delta_{2}^{\mathcal{T}_3} \!\rightarrow\! \Delta_{3}^{\mathcal{T}_3}$} & \textbf{AP}    & $\textbf{49.9}_{\pm3.5}$ &  $49.4_{\pm3.2}$     & $49.9_{\pm3.8}$ & $49.2_{\pm1.2}$ \\
          & \textbf{FT}    & $54.4_{\pm1.2}$ &   $49.4_{\pm3.6}$    & $50.6_{\pm3.0}$ & $\textbf{58.0}_{\pm3.4}$ \\
    \bottomrule
    \end{tabular}%
  \caption{Experiments of distillation-based recyclable tuning. For each task $\mathcal{T}_i$, we evaluate two settings $\Delta_{i}^{\mathcal{T}_i} \!\rightarrow\! \Delta_{i+1}^{\mathcal{T}_i}$ and $\Delta_{i-1}^{\mathcal{T}_i} \!\rightarrow\! \Delta_{i}^{\mathcal{T}_i}$. For example, $\Delta_{i}^{\mathcal{T}_i} \!\rightarrow\! \Delta_{i+1}^{\mathcal{T}_i}$ denotes distilling the knowledge of $\mathcal{M}_{i}$ to $\mathcal{M}_{i+1}$ for task $\mathcal{T}_i$. Here \textbf{AP} / \textbf{FT} refers to adapter / fine-tuning.}
  \label{tab:fewshot_old_task}%
\end{table}%

(b) Second, we evaluate the recyclable tuning from $\mathcal{M}_{i-1}$ to $\mathcal{M}_{i}$, which is denoted as $\Delta_{i-1}^{\mathcal{T}_i} \!\rightarrow\! \Delta_{i}^{\mathcal{T}_i}$. Different from setting (a), here the evaluated task $\mathcal{T}_{i}$ is relevant to the pre-training domain $\mathcal{D}_i$ of the newly released PLM $\mathcal{M}_i$. $\mathcal{M}_i$ performs better than $\mathcal{M}_{i-1}$ since $\mathcal{M}_i$ has acquired more knowledge related to $\mathcal{T}_i$ when learning $\mathcal{D}_i$. In light of this, we explore whether $\mathcal{M}_i$ could achieve better performance than $\mathcal{M}_{i-1}$ even when trained with fewer supervised examples. Specifically, the data size of $\mathcal{M}_{i-1}$ is set to $\{32, 256, 32\}$-shot for $\{\mathcal{T}_1, \mathcal{T}_2, \mathcal{T}_3\}$, and the data size of $\mathcal{M}_{i}$ is set to $\{16, 32, 16\}$-shot, respectively. We also evaluate our method under the zero-shot setting in \cref{sec:distil_zero_shot}.

We compare our method with utilizing only the task loss ($\mathcal{L}_\text{final}$-$\mathcal{L}_\text{KD}$) to validate the benefits of knowledge distillation. Further, we explore combining both distillation-based and initialization-based recyclable tuning ($\mathcal{L}_\text{final}$+\textit{Init.}). This is implemented by first using the outdated weights as the initialization then tuning with $\mathcal{L}_\text{final}$. We also report teacher performance (\textit{Teacher}) as a reference.

\paragraph{Results.} It can be concluded from Table~\ref{tab:fewshot_old_task} that: (1) compared with optimizing only the task loss ($\mathcal{L}_\text{final}$-$\mathcal{L}_\text{KD}$), distilling knowledge from the outdated weights ($\mathcal{L}_\text{final}$) significantly improves the performance, which shows that \textbf{knowledge distillation is an effective way for recyclable tuning}. (2) In general, $\mathcal{L}_\text{final}$+\textit{Init.} leads to better performance than $\mathcal{L}_\text{final}$. This finding reveals that \textbf{both distillation-based and initialization-based methods are complementary to each other} and can be further combined to fully exploit the knowledge in outdated weights. (3) In Table~\ref{tab:fewshot_old_task} setting (a), $\mathcal{M}_{i+1}$ performs worse than $\mathcal{M}_{i}$ on task $\mathcal{T}_i$, which is because $\mathcal{M}_{i+1}$ forgets some knowledge of domain $\mathcal{D}_i$ when learning $\mathcal{D}_{i+1}$. However, such forgetting can be mitigated by designing better continual pre-training algorithms~\citep{qin-etal-2022-elle}. (4) In Table~\ref{tab:fewshot_old_task} setting (b), $\mathcal{M}_{i}$ outperforms $\mathcal{M}_{i-1}$ despite being trained with fewer examples. This shows that the newly acquired knowledge on domain $\mathcal{D}_i$ conduces to $\mathcal{M}_{i}$'s performance in $\mathcal{D}_i$'s relevant task $\mathcal{T}_{i}$, and improves the data efficiency. We further discuss the difference between distillation-based and initialization-based methods in \cref{sec:compare_init_distil}.
\section{Discussion}
\label{sec:discussion}
\paragraph{Training-free Weight Recycling.}
Both methods proposed in \cref{sec:application} necessitate tuning the upgraded PLM. Such a process often relies on abundant computational costs and may be infeasible practically. Given the close connections among continually pre-trained PLMs, we contend that weight recycling can be realized without training. As a preliminary exploration, we show in \cref{sec:projection_learning} that it is possible to learn a cross-task generalizable projection to directly upgrade the outdated weights and make them compatible with the new PLM. Upgrading outdated weights using such a projection requires far fewer computations ($< 0.002\text{\textperthousand}$) and still achieves satisfactory performance.

\paragraph{Downstream-compatible Continual Pre-training.}
From another angle, recyclable tuning addresses the incompatibility between outdated adapted weights and the upgraded PLM from the customer perspective, analogous to the concept of \textit{forward compatibility} in software engineering. In fact, the responsibility for maintaining compatibility can also be shifted to upstream suppliers during PLM upgrading (i.e., \textit{backward compatibility}). Potential solutions include adding regularization terms during continual pre-training to maintain compatibility with existing adapted weights. In this way, we solve the incompatibility problem once and for all, which is more customer-friendly. However, modifying pre-training objectives may come at the cost of reduced model performance.

\paragraph{Broader Application Scenarios.}
Although we primarily focus on recyclable tuning for one specific scenario (i.e., continual pre-training), PLMs may be subject to various types of evolution in practice. For instance, the expansion of model size (e.g., from $\text{T5}_\texttt{BASE}$~\citep{2020t5} to $\text{T5}_\texttt{LARGE}$), the upgrading of model architecture~\citep{chen-etal-2022-bert2bert,lee-thorp-etal-2022-fnet}, the alteration of optimization objective (e.g., from $\text{T5}$ to $\text{T0}$~\citep{sanh2021multitask} and \textsc{UnifiedQA}~\citep{khashabi-etal-2020-unifiedqa}), etc. Once the backbone infrastructure is upgraded, massive adapted weights would become outdated and potentially wasted. Hence we believe recyclable tuning in fact has broader application scenarios and we hope our findings and solutions could inspire more future research in this area.

\section{Conclusion}
In this paper, we formulate the task of recyclable tuning for continual pre-training. We conduct empirical analyses for this task through the lens of model compatibility, linear mode connectivity, and functional similarity. Inspired by the corresponding findings, we explore the practical benefits of recyclable tuning through parameter initialization and knowledge distillation. We also envision our setup to serve as the testbed for other topics, e.g., cross-model knowledge transfer and continual learning.

\section*{Acknowledgments}

This work is supported by the National Key R\&D Program of China (No. 2020AAA0106502), Institute Guo Qiang at Tsinghua University, Beijing Academy of Artificial Intelligence (BAAI).

Yujia Qin and Cheng Qian designed the methods. Yujia Qin wrote the paper. Cheng Qian conducted the experiments. Yankai Lin, Zhiyuan Liu, Maosong Sun, and Jie Zhou advised the project. All authors participated in the discussion.

\section*{Limitations}
We only experiment with two kinds of PLMs ($\text{RoBERTa}_\texttt{BASE}$ and $\text{RoBERTa}_\texttt{LARGE}$ (\cref{sec:additional_exp_initialization} and \cref{sec:roberta_large_distil})), leaving more diverse kinds of PLMs unexplored. While this allows us to demonstrate the effectiveness of our approach on these specific PLMs, it is important for future work to extend our problem setup to a wider range of PLMs in order to fully understand the generalizability of our findings.

\section*{Ethical Statement}
In this research, we consider the following ethical issues:

\begin{itemize} [topsep=1pt, partopsep=1pt, leftmargin=12pt, itemsep=-3pt]
    \item \textbf{Privacy.} Outdated adapted weights may contain information about the data and tasks they were trained on. Thus it is important to consider the potential privacy implications when recycling these weights. Efforts should be taken to ensure that personal or sensitive information is not disclosed during weight recycling.
    \item \textbf{Fairness.} It is crucial to guarantee that the recycling of adapted weights does not introduce biases or unfairly advantage certain tasks or domains. Thorough analysis and testing are needed to make sure that recyclable tuning does not perpetuate or amplify existing inequalities.
    \item \textbf{Responsible AI.} The responsible development and deployment of AI systems require considering the potential impacts on the environment. By improving the efficiency and sustainability of PLM adaptation, recyclable tuning contributes to the responsible development of AI systems.
    \item \textbf{Transparency.} To facilitate the responsible and ethical use of recyclable tuning, it is vital to be transparent about the methods and assumptions underlying them. We encourage future works to clearly document the conditions under which recyclable tuning is effective, as well as the potential limitations or risks.
\end{itemize}
\bibliography{anthology,custom}

\begin{thebibliography}{60}
\expandafter\ifx\csname natexlab\endcsname\relax\def\natexlab#1{#1}\fi

\bibitem[{Aghajanyan et~al.(2021)Aghajanyan, Gupta, Shrivastava, Chen,
  Zettlemoyer, and Gupta}]{aghajanyan-etal-2021-muppet}
Armen Aghajanyan, Anchit Gupta, Akshat Shrivastava, Xilun Chen, Luke
  Zettlemoyer, and Sonal Gupta. 2021.
\newblock \href {https://doi.org/10.18653/v1/2021.emnlp-main.468} {Muppet:
  Massive multi-task representations with pre-finetuning}.
\newblock In \emph{Proceedings of the 2021 Conference on Empirical Methods in
  Natural Language Processing}, pages 5799--5811, Online and Punta Cana,
  Dominican Republic. Association for Computational Linguistics.

\bibitem[{Barbieri et~al.(2020)Barbieri, Camacho-Collados, Espinosa~Anke, and
  Neves}]{barbieri-etal-2020-tweeteval}
Francesco Barbieri, Jose Camacho-Collados, Luis Espinosa~Anke, and Leonardo
  Neves. 2020.
\newblock \href {https://doi.org/10.18653/v1/2020.findings-emnlp.148}
  {{T}weet{E}val: Unified benchmark and comparative evaluation for tweet
  classification}.
\newblock In \emph{Findings of the Association for Computational Linguistics:
  EMNLP 2020}, pages 1644--1650, Online. Association for Computational
  Linguistics.

\bibitem[{Bommasani et~al.(2021)Bommasani, Hudson, Adeli, Altman, Arora, von
  Arx, Bernstein, Bohg, Bosselut, Brunskill
  et~al.}]{bommasani2021opportunities}
Rishi Bommasani, Drew~A Hudson, Ehsan Adeli, Russ Altman, Simran Arora, Sydney
  von Arx, Michael~S Bernstein, Jeannette Bohg, Antoine Bosselut, Emma
  Brunskill, et~al. 2021.
\newblock \href {https://arxiv.org/pdf/2108.07258.pdf} {On the opportunities
  and risks of foundation models}.
\newblock \emph{arXiv preprint arXiv:2108.07258}.

\bibitem[{Brown et~al.(2020)Brown, Mann, Ryder, Subbiah, Kaplan, Dhariwal,
  Neelakantan, Shyam, Sastry, Askell, Agarwal, Herbert{-}Voss, Krueger,
  Henighan, Child, Ramesh, Ziegler, Wu, Winter, Hesse, Chen, Sigler, Litwin,
  Gray, Chess, Clark, Berner, McCandlish, Radford, Sutskever, and
  Amodei}]{NEURIPS2020_1457c0d6}
Tom~B. Brown, Benjamin Mann, Nick Ryder, Melanie Subbiah, Jared Kaplan,
  Prafulla Dhariwal, Arvind Neelakantan, Pranav Shyam, Girish Sastry, Amanda
  Askell, Sandhini Agarwal, Ariel Herbert{-}Voss, Gretchen Krueger, Tom
  Henighan, Rewon Child, Aditya Ramesh, Daniel~M. Ziegler, Jeffrey Wu, Clemens
  Winter, Christopher Hesse, Mark Chen, Eric Sigler, Mateusz Litwin, Scott
  Gray, Benjamin Chess, Jack Clark, Christopher Berner, Sam McCandlish, Alec
  Radford, Ilya Sutskever, and Dario Amodei. 2020.
\newblock \href
  {https://proceedings.neurips.cc/paper/2020/hash/1457c0d6bfcb4967418bfb8ac142f64a-Abstract.html}
  {Language models are few-shot learners}.
\newblock In \emph{Advances in Neural Information Processing Systems 33: Annual
  Conference on Neural Information Processing Systems 2020, NeurIPS 2020,
  December 6-12, 2020, virtual}.

\bibitem[{Chen et~al.(2022)Chen, Yin, Shang, Jiang, Qin, Wang, Wang, Chen, Liu,
  and Liu}]{chen-etal-2022-bert2bert}
Cheng Chen, Yichun Yin, Lifeng Shang, Xin Jiang, Yujia Qin, Fengyu Wang, Zhi
  Wang, Xiao Chen, Zhiyuan Liu, and Qun Liu. 2022.
\newblock \href {https://doi.org/10.18653/v1/2022.acl-long.151} {bert2{BERT}:
  Towards reusable pretrained language models}.
\newblock In \emph{Proceedings of the 60th Annual Meeting of the Association
  for Computational Linguistics (Volume 1: Long Papers)}, pages 2134--2148,
  Dublin, Ireland. Association for Computational Linguistics.

\bibitem[{Davidson et~al.(2017)Davidson, Warmsley, Macy, and
  Weber}]{hateoffensive}
Thomas Davidson, Dana Warmsley, Michael Macy, and Ingmar Weber. 2017.
\newblock \href {https://arxiv.org/pdf/1703.04009.pdf} {Automated hate speech
  detection and the problem of offensive language}.
\newblock In \emph{Proceedings of the 11th International AAAI Conference on Web
  and Social Media}, ICWSM '17, pages 512--515.

\bibitem[{Devlin et~al.(2019)Devlin, Chang, Lee, and
  Toutanova}]{devlin2018bert}
Jacob Devlin, Ming-Wei Chang, Kenton Lee, and Kristina Toutanova. 2019.
\newblock \href {https://doi.org/10.18653/v1/N19-1423} {{BERT}: Pre-training of
  deep bidirectional transformers for language understanding}.
\newblock In \emph{Proceedings of the 2019 Conference of the North {A}merican
  Chapter of the Association for Computational Linguistics: Human Language
  Technologies, Volume 1 (Long and Short Papers)}, pages 4171--4186,
  Minneapolis, Minnesota. Association for Computational Linguistics.

\bibitem[{Ding et~al.(2022)Ding, Qin, Yang, Wei, Yang, Su, Hu, Chen, Chan, Chen
  et~al.}]{ding2022delta}
Ning Ding, Yujia Qin, Guang Yang, Fuchao Wei, Zonghan Yang, Yusheng Su,
  Shengding Hu, Yulin Chen, Chi-Min Chan, Weize Chen, et~al. 2022.
\newblock \href {https://arxiv.org/pdf/2203.06904.pdf} {Delta tuning: A
  comprehensive study of parameter efficient methods for pre-trained language
  models}.
\newblock \emph{arXiv preprint arXiv:2203.06904}.

\bibitem[{Draxler et~al.(2018)Draxler, Veschgini, Salmhofer, and
  Hamprecht}]{draxler2018essentially}
Felix Draxler, Kambis Veschgini, Manfred Salmhofer, and Fred~A. Hamprecht.
  2018.
\newblock \href {http://proceedings.mlr.press/v80/draxler18a.html} {Essentially
  no barriers in neural network energy landscape}.
\newblock In \emph{Proceedings of the 35th International Conference on Machine
  Learning, {ICML} 2018, Stockholmsm{\"{a}}ssan, Stockholm, Sweden, July 10-15,
  2018}, volume~80 of \emph{Proceedings of Machine Learning Research}, pages
  1308--1317. {PMLR}.

\bibitem[{Faruqui and Das(2018)}]{faruqui-das-2018-identifying}
Manaal Faruqui and Dipanjan Das. 2018.
\newblock \href {https://doi.org/10.18653/v1/D18-1091} {Identifying well-formed
  natural language questions}.
\newblock In \emph{Proceedings of the 2018 Conference on Empirical Methods in
  Natural Language Processing}, pages 798--803, Brussels, Belgium. Association
  for Computational Linguistics.

\bibitem[{Frankle et~al.(2020)Frankle, Dziugaite, Roy, and
  Carbin}]{frankle2020linear}
Jonathan Frankle, Gintare~Karolina Dziugaite, Daniel Roy, and Michael Carbin.
  2020.
\newblock \href {http://proceedings.mlr.press/v119/frankle20a.html} {Linear
  mode connectivity and the lottery ticket hypothesis}.
\newblock In \emph{Proceedings of the 37th International Conference on Machine
  Learning, {ICML} 2020, 13-18 July 2020, Virtual Event}, volume 119 of
  \emph{Proceedings of Machine Learning Research}, pages 3259--3269. {PMLR}.

\bibitem[{Freeman and Bruna(2017)}]{freeman2016topology}
C.~Daniel Freeman and Joan Bruna. 2017.
\newblock \href {https://openreview.net/forum?id=Bk0FWVcgx} {Topology and
  geometry of half-rectified network optimization}.
\newblock In \emph{5th International Conference on Learning Representations,
  {ICLR} 2017, Toulon, France, April 24-26, 2017, Conference Track
  Proceedings}. OpenReview.net.

\bibitem[{Garipov et~al.(2018)Garipov, Izmailov, Podoprikhin, Vetrov, and
  Wilson}]{garipov2018loss}
Timur Garipov, Pavel Izmailov, Dmitrii Podoprikhin, Dmitry~P. Vetrov, and
  Andrew~Gordon Wilson. 2018.
\newblock \href
  {https://proceedings.neurips.cc/paper/2018/hash/be3087e74e9100d4bc4c6268cdbe8456-Abstract.html}
  {Loss surfaces, mode connectivity, and fast ensembling of dnns}.
\newblock In \emph{Advances in Neural Information Processing Systems 31: Annual
  Conference on Neural Information Processing Systems 2018, NeurIPS 2018,
  December 3-8, 2018, Montr{\'{e}}al, Canada}, pages 8803--8812.

\bibitem[{Gong et~al.(2019)Gong, He, Li, Qin, Wang, and Liu}]{pmlr-v97-gong19a}
Linyuan Gong, Di~He, Zhuohan Li, Tao Qin, Liwei Wang, and Tieyan Liu. 2019.
\newblock \href {https://proceedings.mlr.press/v97/gong19a.html} {Efficient
  training of {BERT} by progressively stacking}.
\newblock In \emph{Proceedings of the 36th International Conference on Machine
  Learning}, volume~97 of \emph{Proceedings of Machine Learning Research},
  pages 2337--2346. PMLR.

\bibitem[{Gururangan et~al.(2020)Gururangan, Marasovi{\'c}, Swayamdipta, Lo,
  Beltagy, Downey, and Smith}]{gururangan2020don}
Suchin Gururangan, Ana Marasovi{\'c}, Swabha Swayamdipta, Kyle Lo, Iz~Beltagy,
  Doug Downey, and Noah~A. Smith. 2020.
\newblock \href {https://doi.org/10.18653/v1/2020.acl-main.740} {Don{'}t stop
  pretraining: Adapt language models to domains and tasks}.
\newblock In \emph{Proceedings of the 58th Annual Meeting of the Association
  for Computational Linguistics}, pages 8342--8360, Online. Association for
  Computational Linguistics.

\bibitem[{He and McAuley(2016)}]{he2016ups}
Ruining He and Julian~J. McAuley. 2016.
\newblock \href {https://doi.org/10.1145/2872427.2883037} {Ups and downs:
  Modeling the visual evolution of fashion trends with one-class collaborative
  filtering}.
\newblock In \emph{Proceedings of the 25th International Conference on World
  Wide Web, {WWW} 2016, Montreal, Canada, April 11 - 15, 2016}, pages 507--517.
  {ACM}.

\bibitem[{Hinton et~al.(2015)Hinton, Vinyals, and Dean}]{hinton2015distilling}
Geoffrey Hinton, Oriol Vinyals, and Jeff Dean. 2015.
\newblock \href {https://arxiv.org/abs/1503.02531} {Distilling the knowledge in
  a neural network}.
\newblock \emph{ArXiv preprint}, abs/1503.02531.

\bibitem[{Houlsby et~al.(2019)Houlsby, Giurgiu, Jastrzebski, Morrone,
  De~Laroussilhe, Gesmundo, Attariyan, and Gelly}]{pmlr-v97-houlsby19a}
Neil Houlsby, Andrei Giurgiu, Stanislaw Jastrzebski, Bruna Morrone, Quentin
  De~Laroussilhe, Andrea Gesmundo, Mona Attariyan, and Sylvain Gelly. 2019.
\newblock \href {https://proceedings.mlr.press/v97/houlsby19a.html}
  {Parameter-efficient transfer learning for {NLP}}.
\newblock In \emph{Proceedings of the 36th International Conference on Machine
  Learning}, volume~97 of \emph{Proceedings of Machine Learning Research},
  pages 2790--2799. PMLR.

\bibitem[{Jang et~al.(2022)Jang, Ye, Lee, Yang, Shin, Han, Kim, and
  Seo}]{jang2022temporalwiki}
Joel Jang, Seonghyeon Ye, Changho Lee, Sohee Yang, Joongbo Shin, Janghoon Han,
  Gyeonghun Kim, and Minjoon Seo. 2022.
\newblock \href {https://arxiv.org/pdf/2204.14211.pdf} {Temporalwiki: A
  lifelong benchmark for training and evaluating ever-evolving language
  models}.

\bibitem[{Jang et~al.(2021)Jang, Ye, Yang, Shin, Han, Kim, Choi, and
  Seo}]{jang2021towards}
Joel Jang, Seonghyeon Ye, Sohee Yang, Joongbo Shin, Janghoon Han, Gyeonghun
  Kim, Stanley~Jungkyu Choi, and Minjoon Seo. 2021.
\newblock \href {https://arxiv.org/abs/2110.03215} {Towards continual knowledge
  learning of language models}.
\newblock \emph{ArXiv preprint}, abs/2110.03215.

\bibitem[{Jin et~al.(2022)Jin, Zhang, Zhu, Xiao, Li, Wei, Arnold, and
  Ren}]{jin-etal-2022-lifelong}
Xisen Jin, Dejiao Zhang, Henghui Zhu, Wei Xiao, Shang-Wen Li, Xiaokai Wei,
  Andrew Arnold, and Xiang Ren. 2022.
\newblock \href {https://doi.org/10.18653/v1/2022.bigscience-1.1} {Lifelong
  pretraining: Continually adapting language models to emerging corpora}.
\newblock In \emph{Proceedings of BigScience Episode {\#}5 -- Workshop on
  Challenges {\&} Perspectives in Creating Large Language Models}, pages 1--16,
  virtual+Dublin. Association for Computational Linguistics.

\bibitem[{Jurgens et~al.(2018)Jurgens, Kumar, Hoover, McFarland, and
  Jurafsky}]{jurgens2018measuring}
David Jurgens, Srijan Kumar, Raine Hoover, Dan McFarland, and Dan Jurafsky.
  2018.
\newblock \href {https://doi.org/10.1162/tacl_a_00028} {Measuring the evolution
  of a scientific field through citation frames}.
\newblock \emph{Transactions of the Association for Computational Linguistics},
  6:391--406.

\bibitem[{Khashabi et~al.(2020)Khashabi, Min, Khot, Sabharwal, Tafjord, Clark,
  and Hajishirzi}]{khashabi-etal-2020-unifiedqa}
Daniel Khashabi, Sewon Min, Tushar Khot, Ashish Sabharwal, Oyvind Tafjord,
  Peter Clark, and Hannaneh Hajishirzi. 2020.
\newblock \href {https://doi.org/10.18653/v1/2020.findings-emnlp.171}
  {{UNIFIEDQA}: Crossing format boundaries with a single {QA} system}.
\newblock In \emph{Findings of the Association for Computational Linguistics:
  EMNLP 2020}, pages 1896--1907, Online. Association for Computational
  Linguistics.

\bibitem[{Kingma and Ba(2015)}]{kingma2014adam}
Diederik~P. Kingma and Jimmy Ba. 2015.
\newblock \href {http://arxiv.org/abs/1412.6980} {Adam: {A} method for
  stochastic optimization}.
\newblock In \emph{3rd International Conference on Learning Representations,
  {ICLR} 2015, San Diego, CA, USA, May 7-9, 2015, Conference Track
  Proceedings}.

\bibitem[{Kringelum et~al.(2016)Kringelum, Kjaerulff, Brunak, Lund, Oprea, and
  Taboureau}]{kringelum2016chemprot}
Jens Kringelum, Sonny~Kim Kjaerulff, S{\o}ren Brunak, Ole Lund, Tudor~I Oprea,
  and Olivier Taboureau. 2016.
\newblock \href {https://pubmed.ncbi.nlm.nih.gov/26876982/} {Chemprot-3.0: a
  global chemical biology diseases mapping}.
\newblock \emph{Database}, 2016.

\bibitem[{Krishna et~al.(2019)Krishna, Tomar, Parikh, Papernot, and
  Iyyer}]{krishna2019thieves}
Kalpesh Krishna, Gaurav~Singh Tomar, Ankur~P Parikh, Nicolas Papernot, and
  Mohit Iyyer. 2019.
\newblock \href {https://arxiv.org/pdf/1910.12366.pdf} {Thieves on sesame
  street! model extraction of bert-based apis}.
\newblock \emph{arXiv preprint arXiv:1910.12366}.

\bibitem[{Lee-Thorp et~al.(2022)Lee-Thorp, Ainslie, Eckstein, and
  Ontanon}]{lee-thorp-etal-2022-fnet}
James Lee-Thorp, Joshua Ainslie, Ilya Eckstein, and Santiago Ontanon. 2022.
\newblock \href {https://doi.org/10.18653/v1/2022.naacl-main.319} {{FN}et:
  Mixing tokens with {F}ourier transforms}.
\newblock In \emph{Proceedings of the 2022 Conference of the North American
  Chapter of the Association for Computational Linguistics: Human Language
  Technologies}, pages 4296--4313, Seattle, United States. Association for
  Computational Linguistics.

\bibitem[{Lin et~al.(2021)Lin, Li, Wang, Li, Du, Xiao, and
  Zhu}]{lin-etal-2021-weight}
Ye~Lin, Yanyang Li, Ziyang Wang, Bei Li, Quan Du, Tong Xiao, and Jingbo Zhu.
  2021.
\newblock \href {https://doi.org/10.18653/v1/2021.acl-long.162} {Weight
  distillation: Transferring the knowledge in neural network parameters}.
\newblock In \emph{Proceedings of the 59th Annual Meeting of the Association
  for Computational Linguistics and the 11th International Joint Conference on
  Natural Language Processing (Volume 1: Long Papers)}, pages 2076--2088,
  Online. Association for Computational Linguistics.

\bibitem[{Liu et~al.(2019)Liu, Ott, Goyal, Du, Joshi, Chen, Levy, Lewis,
  Zettlemoyer, and Stoyanov}]{liu2019roberta}
Yinhan Liu, Myle Ott, Naman Goyal, Jingfei Du, Mandar Joshi, Danqi Chen, Omer
  Levy, Mike Lewis, Luke Zettlemoyer, and Veselin Stoyanov. 2019.
\newblock \href {https://arxiv.org/abs/1907.11692} {{RoBERTa}: A robustly
  optimized {BERT} pretraining approach}.
\newblock \emph{ArXiv preprint}, abs/1907.11692.

\bibitem[{Lo et~al.(2020)Lo, Wang, Neumann, Kinney, and Weld}]{lo2019s2orc}
Kyle Lo, Lucy~Lu Wang, Mark Neumann, Rodney Kinney, and Daniel Weld. 2020.
\newblock \href {https://doi.org/10.18653/v1/2020.acl-main.447} {{S}2{ORC}: The
  semantic scholar open research corpus}.
\newblock In \emph{Proceedings of the 58th Annual Meeting of the Association
  for Computational Linguistics}, pages 4969--4983, Online. Association for
  Computational Linguistics.

\bibitem[{Loshchilov and Hutter(2019)}]{loshchilov2017decoupled}
Ilya Loshchilov and Frank Hutter. 2019.
\newblock \href {https://openreview.net/forum?id=Bkg6RiCqY7} {Decoupled weight
  decay regularization}.
\newblock In \emph{7th International Conference on Learning Representations,
  {ICLR} 2019, New Orleans, LA, USA, May 6-9, 2019}. OpenReview.net.

\bibitem[{Maas et~al.(2011)Maas, Daly, Pham, Huang, Ng, and
  Potts}]{maas-etal-2011-learning}
Andrew~L. Maas, Raymond~E. Daly, Peter~T. Pham, Dan Huang, Andrew~Y. Ng, and
  Christopher Potts. 2011.
\newblock \href {https://www.aclweb.org/anthology/P11-1015} {Learning word
  vectors for sentiment analysis}.
\newblock In \emph{Proceedings of the 49th Annual Meeting of the Association
  for Computational Linguistics: Human Language Technologies}, pages 142--150,
  Portland, Oregon, USA. Association for Computational Linguistics.

\bibitem[{Marelli et~al.(2014)Marelli, Menini, Baroni, Bentivogli, Bernardi,
  and Zamparelli}]{marelli2014sick}
Marco Marelli, Stefano Menini, Marco Baroni, Luisa Bentivogli, Raffaella
  Bernardi, and Roberto Zamparelli. 2014.
\newblock \href
  {http://www.lrec-conf.org/proceedings/lrec2014/pdf/363_Paper.pdf} {A sick
  cure for the evaluation of compositional distributional semantic models}.
\newblock In \emph{Proceedings of the Ninth International Conference on
  Language Resources and Evaluation (LREC'14)}, pages 216--223.

\bibitem[{McAuley and Leskovec(2013)}]{mcauley2013hidden}
Julian McAuley and Jure Leskovec. 2013.
\newblock \href {https://dl.acm.org/doi/abs/10.1145/2507157.2507163} {Hidden
  factors and hidden topics: understanding rating dimensions with review text}.
\newblock In \emph{Proceedings of the 7th ACM conference on Recommender
  systems}, pages 165--172.

\bibitem[{Mirzadeh et~al.(2020)Mirzadeh, Farajtabar, Gorur, Pascanu, and
  Ghasemzadeh}]{mirzadeh2020linear}
Seyed~Iman Mirzadeh, Mehrdad Farajtabar, Dilan Gorur, Razvan Pascanu, and
  Hassan Ghasemzadeh. 2020.
\newblock \href {https://arxiv.org/pdf/2010.04495.pdf} {Linear mode
  connectivity in multitask and continual learning}.
\newblock \emph{arXiv preprint arXiv:2010.04495}.

\bibitem[{Nie et~al.(2020)Nie, Williams, Dinan, Bansal, Weston, and
  Kiela}]{nie-etal-2020-adversarial}
Yixin Nie, Adina Williams, Emily Dinan, Mohit Bansal, Jason Weston, and Douwe
  Kiela. 2020.
\newblock \href {https://doi.org/10.18653/v1/2020.acl-main.441} {Adversarial
  {NLI}: A new benchmark for natural language understanding}.
\newblock In \emph{Proceedings of the 58th Annual Meeting of the Association
  for Computational Linguistics}, pages 4885--4901, Online. Association for
  Computational Linguistics.

\bibitem[{Ott et~al.(2019)Ott, Edunov, Baevski, Fan, Gross, Ng, Grangier, and
  Auli}]{ott2019fairseq}
Myle Ott, Sergey Edunov, Alexei Baevski, Angela Fan, Sam Gross, Nathan Ng,
  David Grangier, and Michael Auli. 2019.
\newblock \href {https://doi.org/10.18653/v1/N19-4009} {fairseq: A fast,
  extensible toolkit for sequence modeling}.
\newblock In \emph{Proceedings of the 2019 Conference of the North {A}merican
  Chapter of the Association for Computational Linguistics (Demonstrations)},
  pages 48--53, Minneapolis, Minnesota. Association for Computational
  Linguistics.

\bibitem[{Pang and Lee(2005)}]{pang-lee-2005-seeing}
Bo~Pang and Lillian Lee. 2005.
\newblock \href {https://doi.org/10.3115/1219840.1219855} {Seeing stars:
  Exploiting class relationships for sentiment categorization with respect to
  rating scales}.
\newblock In \emph{Proceedings of the 43rd Annual Meeting of the Association
  for Computational Linguistics ({ACL}{'}05)}, pages 115--124, Ann Arbor,
  Michigan. Association for Computational Linguistics.

\bibitem[{Paszke et~al.(2019)Paszke, Gross, Massa, Lerer, Bradbury, Chanan,
  Killeen, Lin, Gimelshein, Antiga et~al.}]{paszke2019pytorch}
Adam Paszke, Sam Gross, Francisco Massa, Adam Lerer, James Bradbury, Gregory
  Chanan, Trevor Killeen, Zeming Lin, Natalia Gimelshein, Luca Antiga, et~al.
  2019.
\newblock \href
  {https://proceedings.neurips.cc/paper/2019/hash/bdbca288fee7f92f2bfa9f7012727740-Abstract.html}
  {Pytorch: An imperative style, high-performance deep learning library}.
\newblock \emph{Advances in neural information processing systems}, 32.

\bibitem[{Pfeiffer et~al.(2020)Pfeiffer, R{\"u}ckl{\'e}, Poth, Kamath,
  Vuli{\'c}, Ruder, Cho, and Gurevych}]{pfeiffer2020AdapterHub}
Jonas Pfeiffer, Andreas R{\"u}ckl{\'e}, Clifton Poth, Aishwarya Kamath, Ivan
  Vuli{\'c}, Sebastian Ruder, Kyunghyun Cho, and Iryna Gurevych. 2020.
\newblock \href {https://arxiv.org/pdf/2007.07779.pdf} {Adapterhub: A framework
  for adapting transformers}.
\newblock In \emph{Proceedings of the 2020 Conference on Empirical Methods in
  Natural Language Processing: System Demonstrations}, pages 46--54.

\bibitem[{Poth et~al.(2021)Poth, Pfeiffer, R{\"u}ckl{\'e}, and
  Gurevych}]{poth-etal-2021-pre}
Clifton Poth, Jonas Pfeiffer, Andreas R{\"u}ckl{\'e}, and Iryna Gurevych. 2021.
\newblock \href {https://doi.org/10.18653/v1/2021.emnlp-main.827} {{W}hat to
  pre-train on? {E}fficient intermediate task selection}.
\newblock In \emph{Proceedings of the 2021 Conference on Empirical Methods in
  Natural Language Processing}, pages 10585--10605, Online and Punta Cana,
  Dominican Republic. Association for Computational Linguistics.

\bibitem[{Qin et~al.(2022{\natexlab{a}})Qin, Lin, Yi, Zhang, Han, Zhang, Su,
  Liu, Li, Sun, and Zhou}]{qin-etal-2022-knowledge}
Yujia Qin, Yankai Lin, Jing Yi, Jiajie Zhang, Xu~Han, Zhengyan Zhang, Yusheng
  Su, Zhiyuan Liu, Peng Li, Maosong Sun, and Jie Zhou. 2022{\natexlab{a}}.
\newblock \href {https://doi.org/10.18653/v1/2022.naacl-main.288} {Knowledge
  inheritance for pre-trained language models}.
\newblock In \emph{Proceedings of the 2022 Conference of the North American
  Chapter of the Association for Computational Linguistics: Human Language
  Technologies}, pages 3921--3937, Seattle, United States. Association for
  Computational Linguistics.

\bibitem[{Qin et~al.(2022{\natexlab{b}})Qin, Qian, Yi, Chen, Lin, Han, Liu,
  Sun, and Zhou}]{qin2022exploring}
Yujia Qin, Cheng Qian, Jing Yi, Weize Chen, Yankai Lin, Xu~Han, Zhiyuan Liu,
  Maosong Sun, and Jie Zhou. 2022{\natexlab{b}}.
\newblock \href {https://arxiv.org/pdf/2210.14102.pdf} {Exploring mode
  connectivity for pre-trained language models}.
\newblock \emph{arXiv preprint arXiv:2210.14102}.

\bibitem[{Qin et~al.(2021)Qin, Wang, Su, Lin, Ding, Liu, Li, Hou, Li, Sun
  et~al.}]{qin2021exploring}
Yujia Qin, Xiaozhi Wang, Yusheng Su, Yankai Lin, Ning Ding, Zhiyuan Liu, Juanzi
  Li, Lei Hou, Peng Li, Maosong Sun, et~al. 2021.
\newblock \href {https://arxiv.org/pdf/2110.07867.pdf} {Exploring
  low-dimensional intrinsic task subspace via prompt tuning}.
\newblock \emph{arXiv preprint arXiv:2110.07867}.

\bibitem[{Qin et~al.(2022{\natexlab{c}})Qin, Zhang, Lin, Liu, Li, Sun, and
  Zhou}]{qin-etal-2022-elle}
Yujia Qin, Jiajie Zhang, Yankai Lin, Zhiyuan Liu, Peng Li, Maosong Sun, and Jie
  Zhou. 2022{\natexlab{c}}.
\newblock \href {https://doi.org/10.18653/v1/2022.findings-acl.220} {{ELLE}:
  Efficient lifelong pre-training for emerging data}.
\newblock In \emph{Findings of the Association for Computational Linguistics:
  ACL 2022}, pages 2789--2810, Dublin, Ireland. Association for Computational
  Linguistics.

\bibitem[{Raffel et~al.(2020)Raffel, Shazeer, Roberts, Lee, Narang, Matena,
  Zhou, Li, and Liu}]{2020t5}
Colin Raffel, Noam Shazeer, Adam Roberts, Katherine Lee, Sharan Narang, Michael
  Matena, Yanqi Zhou, Wei Li, and Peter~J. Liu. 2020.
\newblock \href {http://jmlr.org/papers/v21/20-074.html} {Exploring the limits
  of transfer learning with a unified text-to-text transformer}.
\newblock \emph{Journal of Machine Learning Research}, 21(140):1--67.

\bibitem[{Rajpurkar et~al.(2016)Rajpurkar, Zhang, Lopyrev, and
  Liang}]{rajpurkar-etal-2016-squad}
Pranav Rajpurkar, Jian Zhang, Konstantin Lopyrev, and Percy Liang. 2016.
\newblock \href {https://doi.org/10.18653/v1/D16-1264} {{SQ}u{AD}: 100,000+
  questions for machine comprehension of text}.
\newblock In \emph{Proceedings of the 2016 Conference on Empirical Methods in
  Natural Language Processing}, pages 2383--2392, Austin, Texas. Association
  for Computational Linguistics.

\bibitem[{Sanh et~al.(2021)Sanh, Webson, Raffel, Bach, Sutawika, Alyafeai,
  Chaffin, Stiegler, Scao, Raja et~al.}]{sanh2021multitask}
Victor Sanh, Albert Webson, Colin Raffel, Stephen~H Bach, Lintang Sutawika,
  Zaid Alyafeai, Antoine Chaffin, Arnaud Stiegler, Teven~Le Scao, Arun Raja,
  et~al. 2021.
\newblock \href {https://arxiv.org/pdf/2110.08207.pdf} {Multitask prompted
  training enables zero-shot task generalization}.
\newblock \emph{arXiv preprint arXiv:2110.08207}.

\bibitem[{Schick and Sch{\"u}tze(2021)}]{schick-schutze-2021-exploiting}
Timo Schick and Hinrich Sch{\"u}tze. 2021.
\newblock \href {https://aclanthology.org/2021.eacl-main.20} {Exploiting
  cloze-questions for few-shot text classification and natural language
  inference}.
\newblock In \emph{Proceedings of the 16th Conference of the European Chapter
  of the Association for Computational Linguistics: Main Volume}, pages
  255--269, Online. Association for Computational Linguistics.

\bibitem[{Socher et~al.(2013)Socher, Perelygin, Wu, Chuang, Manning, Ng, and
  Potts}]{socher-etal-2013-recursive}
Richard Socher, Alex Perelygin, Jean Wu, Jason Chuang, Christopher~D. Manning,
  Andrew Ng, and Christopher Potts. 2013.
\newblock \href {https://aclanthology.org/D13-1170} {Recursive deep models for
  semantic compositionality over a sentiment treebank}.
\newblock In \emph{Proceedings of the 2013 Conference on Empirical Methods in
  Natural Language Processing}, pages 1631--1642, Seattle, Washington, USA.
  Association for Computational Linguistics.

\bibitem[{Su et~al.(2022)Su, Wang, Qin, Chan, Lin, Wang, Wen, Liu, Li, Li, Hou,
  Sun, and Zhou}]{su-etal-2022-transferability}
Yusheng Su, Xiaozhi Wang, Yujia Qin, Chi-Min Chan, Yankai Lin, Huadong Wang,
  Kaiyue Wen, Zhiyuan Liu, Peng Li, Juanzi Li, Lei Hou, Maosong Sun, and Jie
  Zhou. 2022.
\newblock \href {https://doi.org/10.18653/v1/2022.naacl-main.290} {On
  transferability of prompt tuning for natural language processing}.
\newblock In \emph{Proceedings of the 2022 Conference of the North American
  Chapter of the Association for Computational Linguistics: Human Language
  Technologies}, pages 3949--3969, Seattle, United States. Association for
  Computational Linguistics.

\bibitem[{Sun et~al.(2019)Sun, Cheng, Gan, and Liu}]{sun-etal-2019-patient}
Siqi Sun, Yu~Cheng, Zhe Gan, and Jingjing Liu. 2019.
\newblock \href {https://doi.org/10.18653/v1/D19-1441} {Patient knowledge
  distillation for {BERT} model compression}.
\newblock In \emph{Proceedings of the 2019 Conference on Empirical Methods in
  Natural Language Processing and the 9th International Joint Conference on
  Natural Language Processing (EMNLP-IJCNLP)}, pages 4323--4332, Hong Kong,
  China. Association for Computational Linguistics.

\bibitem[{Vaswani et~al.(2017)Vaswani, Shazeer, Parmar, Uszkoreit, Jones,
  Gomez, Kaiser, and Polosukhin}]{vaswani2017attention}
Ashish Vaswani, Noam Shazeer, Niki Parmar, Jakob Uszkoreit, Llion Jones,
  Aidan~N. Gomez, Lukasz Kaiser, and Illia Polosukhin. 2017.
\newblock \href
  {https://proceedings.neurips.cc/paper/2017/hash/3f5ee243547dee91fbd053c1c4a845aa-Abstract.html}
  {Attention is all you need}.
\newblock In \emph{Advances in Neural Information Processing Systems 30: Annual
  Conference on Neural Information Processing Systems 2017, December 4-9, 2017,
  Long Beach, CA, {USA}}, pages 5998--6008.

\bibitem[{Vu et~al.(2020)Vu, Wang, Munkhdalai, Sordoni, Trischler,
  Mattarella-Micke, Maji, and Iyyer}]{vu-etal-2020-exploring}
Tu~Vu, Tong Wang, Tsendsuren Munkhdalai, Alessandro Sordoni, Adam Trischler,
  Andrew Mattarella-Micke, Subhransu Maji, and Mohit Iyyer. 2020.
\newblock \href {https://doi.org/10.18653/v1/2020.emnlp-main.635} {Exploring
  and predicting transferability across {NLP} tasks}.
\newblock In \emph{Proceedings of the 2020 Conference on Empirical Methods in
  Natural Language Processing (EMNLP)}, pages 7882--7926, Online. Association
  for Computational Linguistics.

\bibitem[{Williams et~al.(2018)Williams, Nangia, and
  Bowman}]{williams-etal-2018-broad}
Adina Williams, Nikita Nangia, and Samuel Bowman. 2018.
\newblock \href {https://doi.org/10.18653/v1/N18-1101} {A broad-coverage
  challenge corpus for sentence understanding through inference}.
\newblock In \emph{Proceedings of the 2018 Conference of the North {A}merican
  Chapter of the Association for Computational Linguistics: Human Language
  Technologies, Volume 1 (Long Papers)}, pages 1112--1122, New Orleans,
  Louisiana. Association for Computational Linguistics.

\bibitem[{Wolf et~al.(2020)Wolf, Debut, Sanh, Chaumond, Delangue, Moi, Cistac,
  Rault, Louf, Funtowicz, Davison, Shleifer, von Platen, Ma, Jernite, Plu, Xu,
  Scao, Gugger, Drame, Lhoest, and Rush}]{wolf-etal-2020-transformers}
Thomas Wolf, Lysandre Debut, Victor Sanh, Julien Chaumond, Clement Delangue,
  Anthony Moi, Pierric Cistac, Tim Rault, Rémi Louf, Morgan Funtowicz, Joe
  Davison, Sam Shleifer, Patrick von Platen, Clara Ma, Yacine Jernite, Julien
  Plu, Canwen Xu, Teven~Le Scao, Sylvain Gugger, Mariama Drame, Quentin Lhoest,
  and Alexander~M. Rush. 2020.
\newblock \href {https://www.aclweb.org/anthology/2020.emnlp-demos.6}
  {Transformers: State-of-the-art natural language processing}.
\newblock In \emph{Proceedings of the 2020 Conference on Empirical Methods in
  Natural Language Processing: System Demonstrations}, pages 38--45, Online.
  Association for Computational Linguistics.

\bibitem[{Wu et~al.(2021)Wu, Caccia, Li, Li, Qi, and
  Haffari}]{wu2021pretrained}
Tongtong Wu, Massimo Caccia, Zhuang Li, Yuan-Fang Li, Guilin Qi, and Gholamreza
  Haffari. 2021.
\newblock \href {https://openreview.net/pdf?id=figzpGMrdD} {Pretrained language
  model in continual learning: A comparative study}.
\newblock In \emph{International Conference on Learning Representations}.

\bibitem[{Yi et~al.(2022)Yi, Chen, Qin, Lin, Ding, Han, Liu, Sun, and
  Zhou}]{yi-etal-2022-different}
Jing Yi, Weize Chen, Yujia Qin, Yankai Lin, Ning Ding, Xu~Han, Zhiyuan Liu,
  Maosong Sun, and Jie Zhou. 2022.
\newblock \href {https://aclanthology.org/2022.findings-emnlp.244} {Different
  tunes played with equal skill: Exploring a unified optimization subspace for
  parameter-efficient tuning}.
\newblock In \emph{Findings of the Association for Computational Linguistics:
  EMNLP 2022}, pages 3348--3366, Abu Dhabi, United Arab Emirates. Association
  for Computational Linguistics.

\bibitem[{Zellers et~al.(2019)Zellers, Holtzman, Rashkin, Bisk, Farhadi,
  Roesner, and Choi}]{zellers2019defending}
Rowan Zellers, Ari Holtzman, Hannah Rashkin, Yonatan Bisk, Ali Farhadi,
  Franziska Roesner, and Yejin Choi. 2019.
\newblock \href
  {https://proceedings.neurips.cc/paper/2019/hash/3e9f0fc9b2f89e043bc6233994dfcf76-Abstract.html}
  {Defending against neural fake news}.
\newblock In \emph{Advances in Neural Information Processing Systems 32: Annual
  Conference on Neural Information Processing Systems 2019, NeurIPS 2019,
  December 8-14, 2019, Vancouver, BC, Canada}, pages 9051--9062.

\bibitem[{Zhu et~al.(2015)Zhu, Kiros, Zemel, Salakhutdinov, Urtasun, Torralba,
  and Fidler}]{zhu2015aligning}
Yukun Zhu, Ryan Kiros, Richard~S. Zemel, Ruslan Salakhutdinov, Raquel Urtasun,
  Antonio Torralba, and Sanja Fidler. 2015.
\newblock \href {https://doi.org/10.1109/ICCV.2015.11} {Aligning books and
  movies: Towards story-like visual explanations by watching movies and reading
  books}.
\newblock In \emph{2015 {IEEE} International Conference on Computer Vision,
  {ICCV} 2015, Santiago, Chile, December 7-13, 2015}, pages 19--27. {IEEE}
  Computer Society.

\end{thebibliography}
\bibliographystyle{acl_natbib}

\clearpage
\appendix

\section*{Appendices}
\label{sec:appendix}

\section{Additional Backgrounds}
\subsection{Parameter-efficient Tuning}
\label{sec:pet}
Conventional downstream adaptation of PLMs involves optimizing all parameters (i.e., fine-tuning), which may cause a heavy burden on the computational infrastructure and storage space. To efficiently utilize the knowledge contained in PLMs, parameter-efficient tuning (PET) is proposed, which optimizes only a few parameters and freezes the majority of parameters~\citep{pmlr-v97-houlsby19a}. Despite extensively reducing the tunable parameters, PET achieves comparable performance to fine-tuning. Besides, due to its lightweight nature, adapted weights produced by PET are easier to train, store, and share among consumers. Thus we deem PET as an essential component in our problem setup. Without loss of generality, we consider a representative PET algorithm, i.e., adapter~\citep{pmlr-v97-houlsby19a} in this paper. Adapter inserts tunable modules into both the feed-forward module and multi-head attention module of each Transformer~\citep{vaswani2017attention} layer.

\subsection{Mode Connectivity}
\label{sec:mode_connectivity}
Mode connectivity measures whether two minima in the parameter space can be connected by a parametric path, where the loss (performance) remains low (high)~\citep{garipov2018loss,freeman2016topology,draxler2018essentially}. Such a property implies that different minima can potentially form a connected manifold in the loss landscape. For two connected minima, we can interpolate them to obtain a series of high-performance solutions. These solutions can be ensembled to achieve performance~\citep{garipov2018loss} that is better than the endpoints.

Prior works in mode connectivity show that under most cases, in neural networks, there exists a \textit{non-linear} low-loss path between different minima. However, only occasionally a \textit{linear} low-loss path could connect different minima. Later works further contend that it is non-trivial if both minima can be connected by a \textit{linear} path~\citep{frankle2020linear,mirzadeh2020linear}. The linearity indicates that both minima may probably lie in the same loss basin~\citep{qin2022exploring}, which is a more favorable property and indicates a closer connection between both minima. In view of this, we focus on analyzing the linear mode connectivity in this paper.

Previous efforts were mainly spent on investigating mode connectivity for non-pre-trained models, until recently, \citet{qin2022exploring} explore such property for PLMs. They focus on tuning \textbf{one static base model} with different adaptation strategies. Differently, we take the first step to explore mode connectivity for \textbf{different backbone models} (continually pre-trained PLMs) and reveal novel insights. Following \citet{qin2022exploring}, we present the results of task performance (e.g., accuracy) to evaluate the mode connectivity in the main paper and also report the results of task loss in \cref{sec:visualization_loss}.

\section{Training-free Weight Recycling}
\label{sec:projection_learning}
Although we have shown that initialization-based recyclable tuning could accelerate the convergence and improve the training efficiency, tuning the upgraded PLM still requires abundant training computations. Especially considering the massive number of tasks to handle, conducting adaptation for all of them whenever the PLM is upgraded is computationally expensive.

In this section, we explore whether we could alleviate the burden of supervised training, and directly upgrade the outdated weights at a small cost. A desired algorithm should consume significantly lower computations than that of training the new PLM from scratch. Meanwhile, this algorithm should achieve satisfactory task performance.

\subsection{Framework}
Inspired by \citet{qin2021exploring,yi-etal-2022-different}, we propose a training-free weight recycling method. Specifically, we learn a cross-task generalizable projection that could directly produce upgraded adapted weights for a specific task, omitting the labor of supervised training. We contend that although there exist massive downstream tasks, a large percentage of them are intrinsically similar and can be categorized into the same task type (e.g., \textit{sentiment analysis}, \textit{question answering}, etc.). Intuitively, the upgrading of a certain task $\mathcal{T}_1$ should provide a referential experience for that of a similar task $\mathcal{T}_2$. In view of this, we propose to make the upgrading process of $\mathcal{T}_1$ recyclable so that the upgrading of $\mathcal{T}_2$ can be achieved efficiently.

\begin{figure}[!t]
\centering
\includegraphics[width=0.46\textwidth]{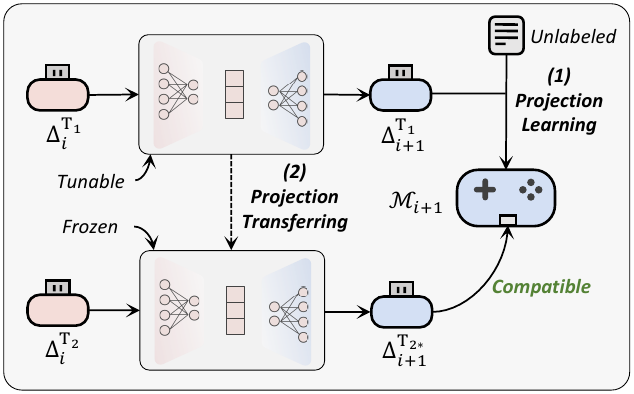}
\caption{Illustration of our training-free weight recycling algorithm. We learn a cross-task generalizable projection (i.e., the projection learning stage) that could directly upgrade outdated adapted weights (i.e., the projection transferring stage).}
\label{fig:method_2}
\end{figure}

For two sequentially released PLMs $\mathcal{M}_i$ and $\mathcal{M}_{i+1}$, assume we have the adapted weights of $\mathcal{M}_i$ for both $\mathcal{T}_1$ and $\mathcal{T}_2$. We aim to recycle these adapted weights for tuning $\mathcal{M}_{i+1}$ on both tasks. As illustrated in Figure~\ref{fig:method_2}, our framework consists of two stages: (1) projection learning and (2) projection transferring. We learn an upgrading projection using task $\mathcal{T}_1$ in the first stage, and then apply (transfer) the learned projection to task $\mathcal{T}_2$ in the second stage. Note the first stage requires training while the second stage is training-free. Next, we introduce the details of the two stages.

\paragraph{Projection Learning.} Instead of directly optimizing the parameters in $\Delta_{i+1}^{\mathcal{T}_1}$, we learn a low-rank decomposition $\Delta_{i+1}^{\mathcal{T}_1} = \texttt{Proj} (\Delta_{i}^{\mathcal{T}_1})$ as follows:
\begin{equation}
\begin{aligned}
   \texttt{Proj}^*_{i \rightarrow i+1} = \argmin_{\texttt{Proj}} \mathcal{L}(\texttt{Proj} (\Delta_{i}^{\mathcal{T}_1})),
    \nonumber
\end{aligned}
\end{equation}
where $\texttt{Proj} \!=\! \texttt{Proj}_{\uparrow} \times \texttt{Proj}_{\downarrow}$. Denote $d$ as a low-dimensional bottleneck dimension, $\texttt{Proj}_{\downarrow}$ projects the dimension of $\Delta_{i}^{\mathcal{T}_1}$ to $d$, i.e., $\texttt{Proj}_{\downarrow}(\Delta_{i}^{\mathcal{T}_1}) \!\in\! \mathbb{R}^{d}$. Then $\texttt{Proj}_{\uparrow}$ projects the dimension from $d$ back to $|\Delta_{i}^{\mathcal{T}_1}|$, i.e., $\texttt{Proj}_{\uparrow}(\texttt{Proj}_{\downarrow}(\Delta_{i}^{\mathcal{T}_1})) \!\in\! \mathbb{R}^{|\Delta_{i}^{\mathcal{T}_1}|}$. Either $\texttt{Proj}_{\uparrow}$ or $\texttt{Proj}_{\downarrow}$ is implemented by a $2$-layer MLP. During training, $\Delta_{i}^{\mathcal{T}_1}$ is kept frozen and only the parameters in $\texttt{Proj}$ are tuned. Note the dimensions of $\Delta_{i}^{\mathcal{T}_1}$ and $\Delta_{i+1}^{\mathcal{T}_1}$ are the same, i.e., $|\Delta_{i}^{\mathcal{T}_1}| \!=\! |\Delta_{i+1}^{\mathcal{T}_1}|$. $\texttt{Proj} (\Delta_{i}^{\mathcal{T}_1})$ is then applied to the upgraded PLM $\mathcal{M}_{i+1}$ to compute the loss $\mathcal{L}$.

\begin{figure}[!t]
\centering
\includegraphics[width=0.46\textwidth]{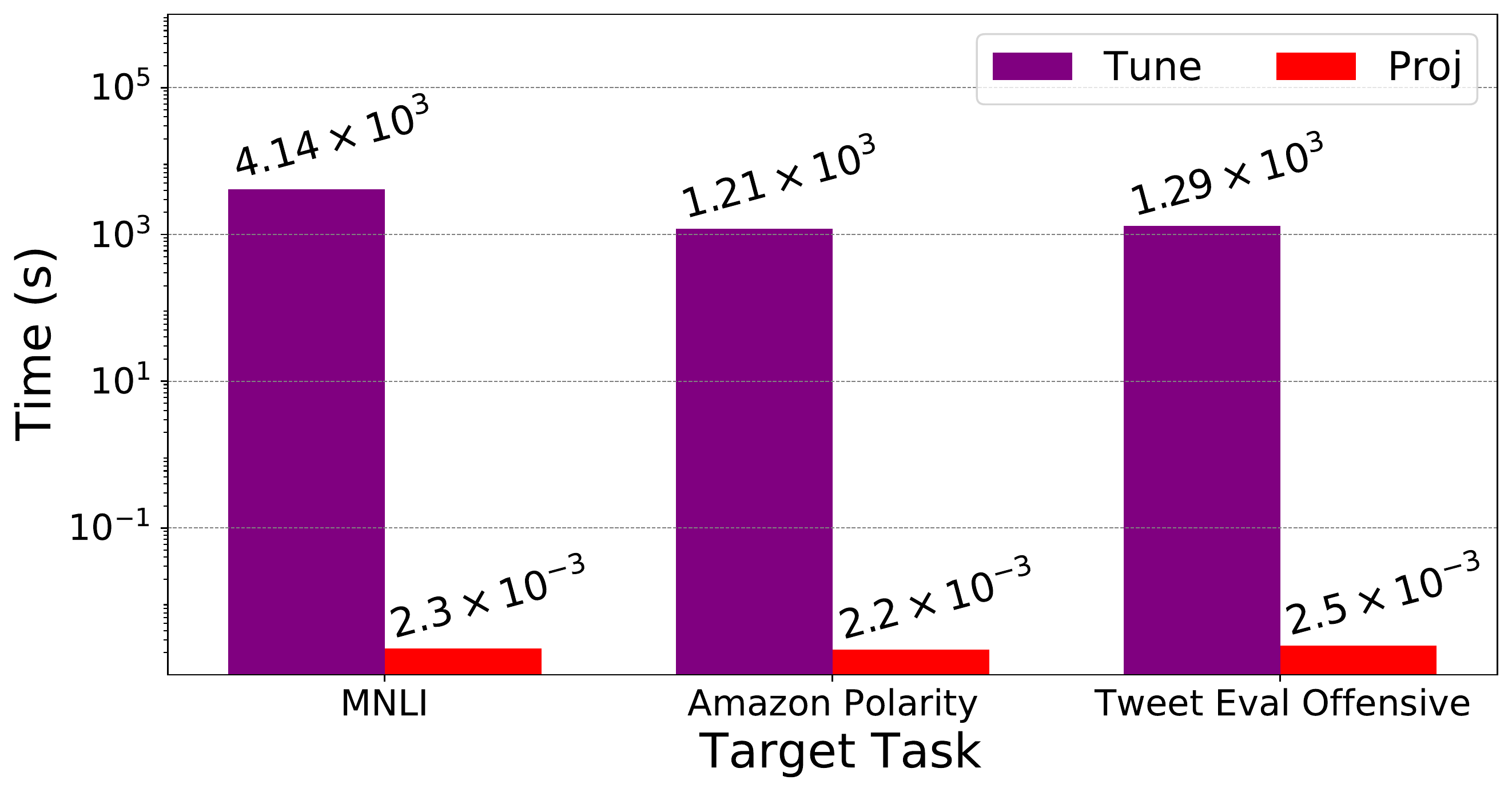}
\caption{Time needed for upgrading outdated weights on different target tasks. Our proposed training-free weight recycling method (Proj) consumes $< 0.002\text{\textperthousand}$ computations than the conventional tuning-based method (Tune).}
\label{fig:training_free_result}
\end{figure}

\paragraph{Projection Transferring.} When upgrading the outdated weights of a similar task $\mathcal{T}_2$, we directly apply the projection $\texttt{Proj}^*_{i \rightarrow i+1}$ learned on $\mathcal{T}_1$ to $\Delta_{i}^{\mathcal{T}_2}$ and obtain the approximated updated weights $\Delta_{i+1}^{\mathcal{T}_{2*}}$:
\begin{equation}
\begin{aligned}
    \Delta_{i+1}^{\mathcal{T}_{2*}} = \texttt{Proj}^*_{i \rightarrow i+1} (\Delta_{i}^{\mathcal{T}_2}).
    \nonumber
\end{aligned}
\end{equation}
We formulate the downstream tuning as \textit{prompt learning}~\citep{schick-schutze-2021-exploiting}, instead of introducing additional classification heads for different tasks. Hence the number of parameters in $\Delta_{i}^{\mathcal{T}_1}$ and $\Delta_{i}^{\mathcal{T}_2}$ is the same, i.e., $|\Delta_{i}^{\mathcal{T}_1}| = |\Delta_{i}^{\mathcal{T}_2}|$. Note that only applying the projection to compute the upgraded weights consumes very limited computations (see Figure~\ref{fig:training_free_result}), hence we significantly reduce the computations of learning $\Delta_{i+1}^{\mathcal{T}_2}$, compared with the conventional tuning-based method.

Besides, since the projection $\texttt{Proj}$ comprises an integral multiple ($d \times$) of $\Delta_{i}^{\mathcal{T}_1}$'s parameters, our solution is only feasible for parameter-efficient tuning. While for fine-tuning, it is computationally intractable to train the projection due to the tremendous size of parameters in $\Delta_{i}^{\mathcal{T}_1}$ and $\texttt{Proj}$. Being the first attempt in this research direction, we leave corresponding explorations for fine-tuning as future work.

\subsection{Experiments}
\label{sec:exp_training}

\paragraph{Settings.}

We mainly evaluate $\mathcal{M}_1$ and $\mathcal{M}_2$ as defined in \cref{sec:task_formulation}. We choose a series of NLP tasks and categorize them into $3$ classes: (1) \textit{natural language inference}: MNLI, SICK, ANLI, QNLI~\citep{rajpurkar-etal-2016-squad}, and WNLI~\citep{faruqui-das-2018-identifying}, (2) \textit{sentiment analysis}: SST-2, Amazon Polarity~\citep{mcauley2013hidden}, and Rotten Tomatoes, (3) \textit{emotion detection}: Hate Speech, Tweet Eval-Offensive, Tweet Eval-Hate, Tweet Eval-Abortion, Tweet Eval-Feminist, and Tweet Eval-Atheism from \citet{barbieri-etal-2020-tweeteval}. We partition the tasks belonging to the same category into source task $\mathcal{T}_1$ and target task $\mathcal{T}_2$ (see Table~\ref{tab:training_efficiency}), and learn the projection $\texttt{Proj}$ on the source task.

We consider the zero-shot setting for the first stage (projection learning) and use the knowledge distillation loss function $\mathcal{L}_\text{KD}$. Here the teacher model weights are the adapted $\mathcal{M}_1$, and the student model weights are obtained by applying $\texttt{Proj} (\Delta_{1}^{\mathcal{T}_1})$ to the pre-trained weights of $\mathcal{M}_{2}$. For the unlabeled corpus used for distillation, we evaluate both the target task data (denoted as $\mathcal{L}_\text{KD}^{\mathcal{T}}$) and Wikipedia corpora ($\mathcal{L}_\text{KD}^{\text{wiki}}$). Note for the former, we only use the input $x$ and discard the corresponding label $y$ (i.e., the zero-shot setting). The former can be seen as the upper bound for the latter since the data format of the latter may not be compatible with the target task. After that, we directly utilize the learned projection to upgrade the outdated weights of similar target tasks.

\paragraph{Baselines.}
We consider \textit{demonstration learning}~\citep{NEURIPS2020_1457c0d6} as the baseline, which integrates a few labeled examples into the input text as additional context. The PLM directly performs inference on the test set without incurring any training. For reference, we also report the performance when $\mathcal{M}_2$ is adapted using the full dataset (FD) and the $32$-shot dataset (FS). Instead, our method requires no labeled data.

\begin{table}[!tbp]
  \centering
  \small
    \begin{tabular}{c@{~}c@{~~}|c@{~~}c@{~}c@{~}c@{~~}c}
    \toprule
    \textbf{Source} & \textbf{Target} & $\mathcal{L}_\text{KD}^{\mathcal{T}}$ & $\mathcal{L}_\text{KD}^{\text{wiki}}$ & Demo. & FD & FS \\
    \midrule
    MNLI & SICK  & $75.2$ & $74.3$ & $60.5$ & \textbf{88.1} & $78.0$ \\
    ANLI  & MNLI & $65.2$ & $43.7$  & $46.1$ & \textbf{79.9} & $41.7$ \\
    QNLI & WNLI & \textbf{72.2} & $58.3$ & $55.6$ & $55.6$ & $50.0$ \\
    \midrule
    SST-2 & A. Polarity & $95.0$ & $94.0$  & $81.8$  & \textbf{95.8} & $94.1$ \\
    SST-2 & R. Tomatoes & \textbf{87.8} & $84.7$ & $71.4$ & $87.4$ & $78.7$ \\
    \midrule
    H. Speech & T. Offensive & $77.1$ & $74.5$ & $63.5$  & \textbf{84.5} & $71.4$ \\
    H. Speech & T. Hate & \textbf{62.4} & $59.1$ & $50.7$ & $52.7$ & $49.2$ \\
    Abortion & Feminist & $63.5$ & \textbf{64.6} & $47.7$ & $59.0$ & $49.5$ \\
    Abortion & Atheism & $71.8$ & $70.0$ & $51.4$ & \textbf{74.6} & $65.0$ \\
    \bottomrule
    \end{tabular}%
  \caption{Performance evaluation for our training-free upgrading algorithm. We distill the knowledge of outdated weights using both unlabeled task data ($\mathcal{L}_\text{KD}^{\mathcal{T}}$) and Wikipedia ($\mathcal{L}_\text{KD}^{\text{wiki}}$). We also report the performance of directly tuning the upgraded PLM using the full dataset (FD) or $32$-shot data (FS), and the demonstration learning baseline (Demo.).}
  \label{tab:training_efficiency}%
\end{table}%

\begin{figure}[!t]
    \centering
    \subfigure{\includegraphics[width=0.45\textwidth]{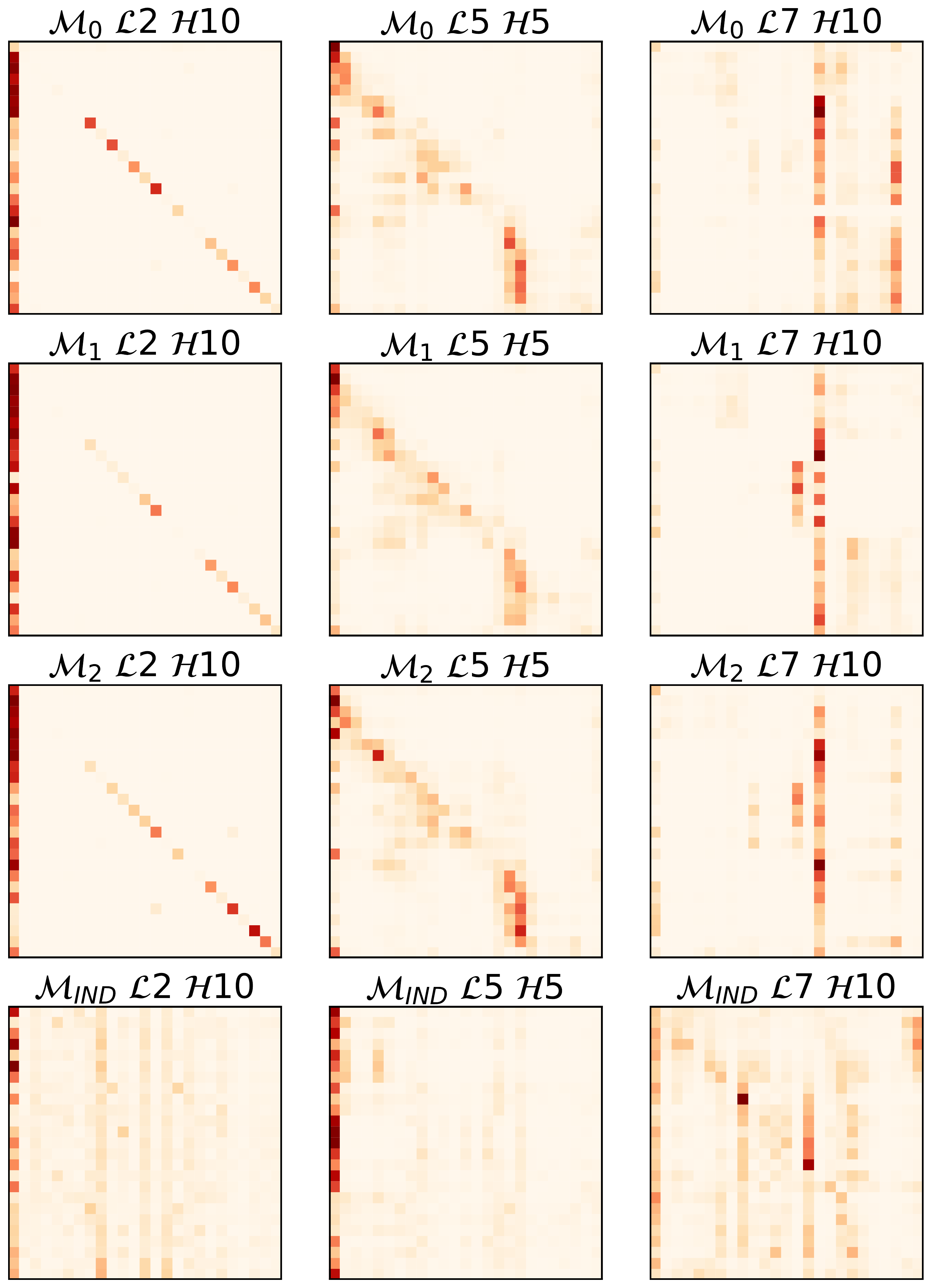}}
    \caption{More attention visualization of attention heads in fine-tuned $\mathcal{M}_0$, $\mathcal{M}_1$, and $\mathcal{M}_2$ given the same input. Apart from the above PLMs, we also present the attention pattern of an independently trained PLM $\mathcal{M}_\text{IND}$.}
    \label{fig:more_attention}
\end{figure}

\begin{figure*}[!t]
\centering
\includegraphics[width=0.85\textwidth]{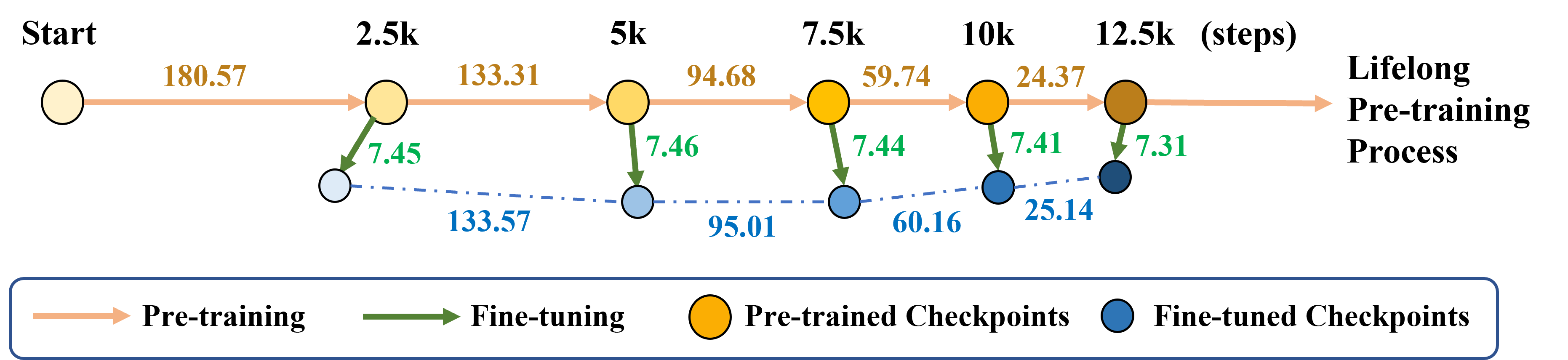}
\caption{Euclidean distance of continually pre-trained PLMs and their adapted weights. During pre-training, we save the checkpoint for every $2.5$k steps and fine-tune each checkpoint on the task \textsc{ChemProt}.}
\label{fig:l2_dist}
\end{figure*}

\paragraph{Efficiency Evaluation.}
We compare the computational costs needed for our training-free method and the conventional tuning-based method in Figure~\ref{fig:training_free_result}. For the former, we record the time needed in projection transferring (i.e., computing the upgraded weights $\Delta_{i+1}^{\mathcal{T}_{2*}}$). For the latter, we record the training time needed until an adaptation converges. It can be derived that our method requires significantly fewer computations, which demonstrates its efficiency. In practice, such a projection can be trained once and for all. As long as we have obtained the projection, we can directly upgrade potentially massive outdated weights in an efficient manner, and the computations involved during projection learning can be neglected. Although currently we only support projection transferring for a similar target task that belongs to the same category of the source task, we expect future work to explore how to train a universal projection that could be applied to an arbitrary task.

\paragraph{Performance Evaluation.}
The results are shown in Table~\ref{tab:training_efficiency}, from which we find that: (1) our method generally outperforms the demonstration baseline, and could surpass the supervised performance (FD and FS) under certain cases, despite not using any labeled data. Hence, besides being computationally efficient, our method achieves satisfactory performance in general. This also validates our intuition that for continually pre-trained PLMs, the upgrading of a specific task could provide referential experience for similar tasks; (2) using the task data ($\mathcal{L}_{\text{KD}}^{\mathcal{T}}$) for distillation generally performs better than using Wikipedia ($\mathcal{L}_{\text{KD}}^{\text{wiki}}$), showing the importance of proper data distribution used for knowledge distillation.

\section{Additional Experiments and Analyses}
\label{sec:additional_exp}

\subsection{Euclidean Distance Analysis}
\label{sec:euclidean_dist}

We report the Euclidean distance of continually pre-trained PLMs and the corresponding adapted models. We evaluate when the official $\text{RoBERTa}_\texttt{BASE}$ is adapted on the \textsc{Bio} domain for $12.5$k steps following the settings in \cref{sec:task_formulation}. We save the checkpoint for every $2.5$k steps. For each checkpoint $\mathcal{M}_1(t)$, denote its weights as $\theta_{1}^0(t)$, we fine-tune it on \textsc{ChemProt} to obtain its adapted weights $\Delta_1^\mathcal{T}(t)$, where $|\Delta_1^{\mathcal{T}}(t)| = |\theta_{1}^0(t)|$. The resultant model weights are $\theta_{1}^\mathcal{T}(t) = \theta_{1}^0(t) \oplus \Delta_1^\mathcal{T}(t)$.

\begin{table}[!t]
  \centering
  \small
    \begin{tabular}{c@{~~}|c@{~~}c@{~~}c@{~~}c@{~~}}
    \toprule
    \textbf{Initialization} & \textit{Random} & $\mathcal{T}_{\text{diff}}$ & $\mathcal{T}_{\text{sim}}$ & $\mathcal{T}_{\text{same}}$ \\
    \midrule
    ANLI  & $47.4_{\pm0.9}$  & $46.3_{\pm1.1}$  &  $\textbf{49.6}_{\pm1.1}$    & $48.8_{\pm0.9}$  \\
    SICK  & $88.8_{\pm0.2}$  & $88.8_{\pm0.3}$  &  $89.4_{\pm0.4}$  & $\textbf{89.5}_{\pm0.2}$  \\
    H. Speech & $79.9_{\pm3.1}$  &   $\textbf{82.4}_{\pm0.9}$    &  $78.4_{\pm2.0}$   & $81.1_{\pm2.6}$  \\
    \midrule
    Avg. & $72.0_{\pm1.4}$  &   $72.5_{\pm0.8}$   &   $72.5_{\pm1.2}$    &  $\textbf{73.1}_{\pm1.2}$ \\
    \bottomrule
    \end{tabular}%
  \caption{The best test performance on $3$ target tasks with fine-tuning from different initialization using $\text{RoBERTa}_\texttt{BASE}$.}
  \label{tab:init_finetune}%
\end{table}

Given two continually pre-trained models $\mathcal{M}_1(t)$ and $\mathcal{M}_1(t')$, where $t' \!=\! t + 2.5\text{k}$, we flatten their pre-trained weights $\theta_{1}^0(t)$ and $\theta_{1}^0(t')$, and calculate their $L$-2 norm\footnote{We use \texttt{torch.dist} function in PyTorch~\citep{paszke2019pytorch} for implementation.}: $||\theta_{1}^0(t') - \theta_{1}^0(t)||$. In addition, we also calculate the $L$-2 norm of flattened adapted weights ($||\Delta_1^\mathcal{T}(t)||$ / $||\Delta_{1}^\mathcal{T}(t')||$) and distance between adapted PLMs ($||\theta_{1}^\mathcal{T}(t') - \theta_{1}^\mathcal{T}(t)||$). We illustrate the results in Figure~\ref{fig:l2_dist} and find that: (1) $||\theta_{1}^0(t') - \theta_{1}^0(t)||$ / $||\theta_{1}^\mathcal{T}(t') - \theta_{1}^\mathcal{T}(t)||$ gradually decreases with $t$ increasing. This is mainly because the learning rates are warmed up for the first $6\%$ steps, and the learning rate starts to decrease at the $0.75$k-th step, which means the PLM gradually moves slower in the parameter space; (2) the parameter change caused by downstream adaptation (i.e., $||\Delta_1^\mathcal{T}(t)||$ / $||\Delta_{1}^\mathcal{T}(t')||$) is far smaller than that brought by continual pre-training ($||\theta_{1}^0(t') - \theta_{1}^0(t)||$). This is because downstream adaptation converges shortly. After convergence, the model parameters generally stay in a specific optimal region. While continual pre-training constantly pushes the model weights away from the previous checkpoints in the parameter space. Another reason is that continual pre-training uses a large batch $2048$, while downstream adaptation often uses a much smaller batch size (e.g., $16$).

\subsection{More Visualization for Functional Similarity Analysis}
\label{sec:more_visualziation_attention}
In the main paper (Figure~\ref{fig:attention}), we visualize three different attention heads of $\mathcal{M}_0$, $\mathcal{M}_1$, and $\mathcal{M}_2$. In this section, we present more visualizations to further support our claim. We also visualize the attention pattern of an independently trained PLM $\mathcal{M}_\text{IND}$. The results in Figure~\ref{fig:more_attention} again demonstrate our claim that continually pre-trained PLMs exhibit similar attention patterns, which independently trained PLMs do not have.

\subsection{Initialization-based Recyclable Tuning for Fine-tuning and $\text{RoBERTa}_\texttt{LARGE}$}
\label{sec:additional_exp_initialization}

In \cref{sec:application_init}, we mainly evaluate initialization-based recyclable tuning using $\text{RoBERTa}_\texttt{BASE}$ and adapter tuning. Here we extend the experiments to either fine-tuning (Table~\ref{tab:init_finetune}) or $\text{RoBERTa}_\texttt{LARGE}$ (Table~\ref{tab:init_adapter_large}). We choose $3$ tasks in Table~\ref{tab:init} and follow most of the settings. From Table~\ref{tab:init_finetune} and Table~\ref{tab:init_adapter_large}, we find that the main conclusions are generally consistent with those mentioned in the main paper. This implies that the initialization-based method can be applied to different tuning methods and PLMs.

\begin{table}[!t]
  \centering
  \small
    \begin{tabular}{c@{~~}|c@{~~}c@{~~}c@{~~}c@{~~}}
    \toprule
    \textbf{Initialization} & \textit{Random} & $\mathcal{T}_{\text{diff}}$ & $\mathcal{T}_{\text{sim}}$ & $\mathcal{T}_{\text{same}}$ \\
    \midrule
    ANLI  & $56.6_{\pm0.3}$  & $57.0_{\pm0.2}$  &  $\textbf{61.0}_{\pm0.4}$    & $59.9_{\pm0.5}$  \\
    SICK  & $89.8_{\pm0.3}$  & $89.6_{\pm0.1}$  &  $\textbf{91.5}_{\pm0.3}$  & $90.6_{\pm0.4}$  \\
    H. Speech & $84.7_{\pm1.3}$  &   $82.1_{\pm0.5}$    &  $83.6_{\pm0.7}$   & $\textbf{85.3}_{\pm0.4}$  \\
    \midrule
    Avg. & $77.0_{\pm0.6}$  &   $76.2_{\pm0.3}$   &   $\textbf{78.7}_{\pm0.5}$    &  $78.6_{\pm0.4}$ \\
    \bottomrule
    \end{tabular}%
  \caption{The best test performance on $3$ target tasks with adapter tuning from different initialization using $\text{RoBERTa}_\texttt{LARGE}$.}
  \label{tab:init_adapter_large}%
\end{table}

\subsection{Distillation-based Recyclable Tuning under the Zero-shot Setting}
\label{sec:distil_zero_shot}

We extend our distillation-based recyclable tuning to the zero-shot setting where there is no labeled data for tuning the upgraded PLM. We show that it is able to utilize unlabeled raw corpora to distill the knowledge of outdated weights. Specifically, we remove the task loss $\mathcal{L}_{\mathcal{T}}$ in $\mathcal{L}_\text{final}$ and only retain $\mathcal{L}_\text{KD}$. Instead of using supervised examples, we sample unlabeled data $x$ from Wikipedia to compute $\mathcal{L}_\text{KD}$. We evaluate recyclable tuning between $\mathcal{M}_1$ and $\mathcal{M}_2$ and choose $4$ downstream tasks, i.e., \textsc{ChemProt}, IMDB, SST-2, and MNLI. For each task, the outdated weights of $\mathcal{M}_1$ are obtained with the full dataset, and our goal is to distill their knowledge and optimize $\mathcal{M}_2$'s weights.

Two training-free baselines are considered: (1) \textit{manual prompting}~\citep{schick-schutze-2021-exploiting}, which restructures the input into templates by inserting prompts, and (2) \textit{demonstration learning}, which has been introduced in \cref{sec:exp_training}. For both baselines, the PLM directly performs inference on the test set without incurring any training. Moreover, we also evaluate the performance when knowledge distillation is combined with the initialization-based method.

\begin{table}[!tbp]
  \centering
  \small
    \begin{tabular}{c@{~}|c@{}c@{}c@{~}c@{~~}c@{~}c@{~~}c@{~}c@{~}}
    \toprule
    \textbf{Task}  & \multicolumn{2}{c}{\textsc{ChemProt}} & \multicolumn{2}{c}{IMDB} & \multicolumn{2}{c}{SST-2} & \multicolumn{2}{c}{MNLI} \\
    \midrule
    Prompt &  \multicolumn{2}{c}{$8.9$} & \multicolumn{2}{c}{$74.4$}  &   \multicolumn{2}{c}{$81.2$}  &  \multicolumn{2}{c}{$44.4$} \\
    Demo. &  \multicolumn{2}{c}{$9.8$} & \multicolumn{2}{c}{$78.1$}  &   \multicolumn{2}{c}{$84.4$}  &  \multicolumn{2}{c}{$47.1$} \\
    \midrule
    \textbf{Method} & AP & FT & AP & FT & AP & FT & AP & FT \\
    \midrule
    $\mathcal{L}_{\text{KD}}$ & $63.8$ & $67.4$ & $89.4$  &  $88.6$  & $90.6$ & $92.0$ & $56.5$ & $78.3$ \\
    $\mathcal{L}_{\text{KD}}$+\textit{Init.} & $\textbf{73.5}$ & $\textbf{76.0}$ & $\textbf{90.3}$ & $\textbf{90.4}$  & $\textbf{92.5}$ & $\textbf{92.5}$ & $\textbf{76.0}$ & $\textbf{78.3}$ \\
    \bottomrule
    \end{tabular}%
  \caption{Zero-shot experiments for distillation-based recyclable tuning between $\mathcal{M}_1$ and $\mathcal{M}_2$. The outdated weights of $\mathcal{M}_1$ are trained using the full dataset. Both manual prompting (Prompt) and demonstration learning (Demo.) are compared as the baseline.}
  \label{tab:zeroshot}%
\end{table}%

We list the results in Table~\ref{tab:zeroshot}, from which it can be derived that: (1) our method surpasses manual prompting and demonstration learning by a large margin, which shows the benefits of recycling outdated adapted weights in the zero-shot setting; (2) initializing tunable weights with the outdated weights could further improve the performance of $\mathcal{L}_\text{KD}$, which again demonstrates that both initialization-based and distillation-based methods are complementary to each other.

\subsection{Interpolation Distillation}
\label{sec:interpolation_distillation}

Traditional knowledge distillation frameworks have no assumptions about the parametric connection between the teacher and the student, and resort to pulling closer their predictions ($\mathcal{P}$) or inner representations ($\mathbf{h}$). As we have shown in the main paper, continually pre-trained PLMs are guaranteed with close parametric connections. Therefore, traditional knowledge distillation methods may fail to exploit the parametric knowledge contained in the teacher model's parameters. Here we explore another way for more effective distillation-based recyclable tuning under our setting.

\begin{table}[!t]
  \centering
  \small
    \begin{tabular}{c@{~}c@{~~}c@{~}c@{~}c@{~}c@{~}}
    \toprule
    \multicolumn{2}{c}{\textbf{Method}} & $\mathcal{L}_\text{final}$-$\mathcal{L}_\text{KD}$  & $\mathcal{L}_\text{final}$ & $\mathcal{L}_\text{final}$+\textit{Init.} & $\mathcal{L}_\text{ITP}$ \\
    \midrule
    \multicolumn{6}{c}{\textit{Setting (a):} $\Delta_{i}^{\mathcal{T}_i} \!\rightarrow\! \Delta_{i+1}^{\mathcal{T}_i}$, $i \in \{1,2,3\}$} \\
    \midrule
    \multirow{2}[2]{*}{$\Delta_{1}^{\mathcal{T}_1} \!\rightarrow\! \Delta_{2}^{\mathcal{T}_1}$} & \textbf{AP}    & $58.0_{\pm0.9}$ & $62.4_{\pm1.3}$ & $63.8_{\pm3.2}$ & $\textbf{64.4}_{\pm1.9}$\\
          & \textbf{FT}  & $61.4_{\pm3.1}$  & $64.5_{\pm0.5}$ & $64.7_{\pm0.6}$ & $\textbf{64.8}_{\pm0.8}$ \\
    \midrule
    \multirow{2}[2]{*}{$\Delta_{2}^{\mathcal{T}_2} \!\rightarrow\! \Delta_{3}^{\mathcal{T}_2}$} & \textbf{AP}    & $78.3_{\pm1.4}$    & $80.7_{\pm0.3}$ & $80.8_{\pm0.7}$ & $\textbf{80.9}_{\pm0.3}$ \\
          & \textbf{FT}    &  $76.7_{\pm2.2}$     & $79.5_{\pm1.5}$ & $79.7_{\pm1.9}$ & $\textbf{80.2}_{\pm0.3}$ \\
    \midrule
    \multirow{2}[2]{*}{$\Delta_{3}^{\mathcal{T}_3} \!\rightarrow\! \Delta_{4}^{\mathcal{T}_3}$} & \textbf{AP}    &  $48.2_{\pm2.9}$     & $48.0_{\pm1.4}$ & $\textbf{55.9}_{\pm3.9}$ & $51.3_{\pm3.2}$ \\
          & \textbf{FT}  & $51.8_{\pm4.2}$    & $54.2_{\pm0.7}$ & $\textbf{61.4}_{\pm2.9}$ & $55.6_{\pm3.2}$ \\
    \midrule
    \multicolumn{6}{c}{\textit{Setting (b):} $\Delta_{i-1}^{\mathcal{T}_i} \!\rightarrow\! \Delta_{i}^{\mathcal{T}_i}$, $i \in \{1,2,3\}$} \\
    \midrule
    \multirow{2}[2]{*}{$\Delta_{0}^{\mathcal{T}_1} \!\rightarrow\! \Delta_{1}^{\mathcal{T}_1}$} & \textbf{AP}    & $53.1_{\pm0.7}$    & $61.4_{\pm1.1}$ & $\textbf{64.7}_{\pm0.4}$ & $62.3_{\pm1.4}$\\
          & \textbf{FT}  & $56.6_{\pm1.2}$    & $59.3_{\pm1.5}$ & $\textbf{63.4}_{\pm0.7}$ & $62.8_{\pm1.2}$ \\
    \midrule
    \multirow{2}[2]{*}{$\Delta_{1}^{\mathcal{T}_2} \!\rightarrow\! \Delta_{2}^{\mathcal{T}_2}$} & \textbf{AP}    & $84.8_{\pm1.3}$ & $86.0_{\pm0.2}$ & $\textbf{87.3}_{\pm0.4}$ & $86.2_{\pm0.4}$ \\
          & \textbf{FT}  &  $82.0_{\pm1.8}$  & $85.5_{\pm0.8}$ & $\textbf{86.8}_{\pm0.7}$ & $85.7_{\pm1.0}$ \\
    \midrule
    \multirow{2}[2]{*}{$\Delta_{2}^{\mathcal{T}_3} \!\rightarrow\! \Delta_{3}^{\mathcal{T}_3}$} & \textbf{AP}    & $49.4_{\pm3.2}$ & $49.9_{\pm3.8}$ & $49.2_{\pm1.2}$ & $\textbf{50.4}_{\pm3.3}$ \\
          & \textbf{FT}  & $49.4_{\pm3.6}$    & $50.6_{\pm3.0}$ & $\textbf{58.0}_{\pm3.4}$ & $86.2_{\pm0.4}$ \\
    \bottomrule
    \end{tabular}%
  \caption{Performance of interpolation distillation. We follow the settings in \cref{sec:application_distil}. The results of $\mathcal{L}_\text{final}$-$\mathcal{L}_\text{KD}$, $\mathcal{L}_\text{final}$, and $\mathcal{L}_\text{final}$+\textit{Init.} are borrowed from Table~\ref{tab:fewshot_old_task}.}
  \label{tab:fewshot_ITP}%
\end{table}%

\paragraph{Framework.} Inspired by MC-SGD~\citep{mirzadeh2020linear}, we propose an interpolation distillation technique to fully exploit the parametric knowledge contained in outdated adapted weights. Specifically, for recyclable tuning between $\mathcal{M}_{i}$ and $\mathcal{M}_{i+1}$, instead of optimizing the overall loss function using the only endpoint checkpoint ($\theta_{i+1}^{\mathcal{L}_j} = \theta_{i+1}^0 \oplus \Delta_{i+1}^{\mathcal{T}_j}$) for task $\mathcal{T}_j$, we linearly interpolate $\theta_{i}^{\mathcal{L}_j}$ and $\theta_{i+1}^{\mathcal{L}_j}$ to obtain a series of model checkpoints: $\theta (\mu) = (1-\mu) \theta_{i}^{\mathcal{L}_j} + \mu \theta_{i+1}^{\mathcal{L}_j}$. After that, we feed data into $\theta (\mu)$ and minimize the corresponding loss together with $\mathcal{L}(\theta_{i+1}^{\mathcal{L}_j})$:
\begin{equation}
\begin{aligned}
    \mathcal{L}_{\text{ITP}}(\Delta_{i+1}^{\mathcal{T}_j}) = \mathcal{L}(\theta_{i+1}^{\mathcal{L}_j}) \!+\! \gamma\!\!\!\!\!\!\sum_{\mu \in \{\frac{1}{N_\mu}, \cdots, \frac{N_\mu-1}{N_\mu}\}}\!\!\!\!\!\! \mathcal{L}(\theta(\mu)),
    \nonumber
\end{aligned}
\end{equation}
where $\gamma$ is a hyper-parameter, and $N_\mu$ denotes a constant integer. In practice, we found a small $N_\mu$ (e.g., $2$) already achieves satisfying performance. During optimization, only $\Delta_{i+1}^{\mathcal{T}_j}$ is tuned by receiving gradients from both $\mathcal{L}(\theta_{i+1}^j)$ and $\mathcal{L}(\theta(\mu))$.

\paragraph{Experiments.}
We follow most of the settings in \cref{sec:application_distil} and evaluate the performance of interpolation distillation. We compare it with the results of $\mathcal{L}_\text{final}$-$\mathcal{L}_\text{KD}$, $\mathcal{L}_\text{final}$, and $\mathcal{L}_\text{final}$+\textit{Init.}. All results are shown in Table~\ref{tab:fewshot_ITP}, from which we observe that the interpolation distillation method ($\mathcal{L}_\text{ITP}$) generally outperforms the vanilla distillation ($\mathcal{L}_\text{final}$), and could surpass $\mathcal{L}_\text{final}$+\textit{Init.} in certain cases. This shows that interpolation distillation successfully exploits the parametric knowledge contained in the outdated adapted weights, and serves as an improved method for the distillation-based method.

\begin{table}[!tbp]
  \centering
  \small
    \begin{tabular}{c@{~}c@{~~}c@{~}c@{~}c@{~}c@{~}}
    \toprule
    \multicolumn{2}{c}{\textbf{Method}} & $\mathcal{L}_\text{final}$-$\mathcal{L}_\text{KD}$  & $\mathcal{L}_\text{final}$ &  $\mathcal{L}_\text{ITP}$ & $\mathcal{L}_\text{final}$+\textit{Init.} \\
    \midrule
    \multicolumn{6}{c}{\textit{Setting: full-data teacher}} \\
    \midrule
    \multirow{2}[2]{*}{$\Delta_{1}^{\mathcal{T}_1} \!\rightarrow\! \Delta_{2}^{\mathcal{T}_1}$} & \textbf{AP}    & $58.0_{\pm0.9}$ &   $71.3_{\pm1.9}$    & $\textbf{76.8}_{\pm0.6}$ & $73.1_{\pm1.3}$ \\
          & \textbf{FT}    & $61.4_{\pm3.1}$ &   $70.7_{\pm1.5}$    & $74.4_{\pm0.8}$ & $\textbf{76.2}_{\pm0.5}$ \\
    \midrule
    \multirow{2}[2]{*}{$\Delta_{2}^{\mathcal{T}_2} \!\rightarrow\! \Delta_{3}^{\mathcal{T}_2}$} & \textbf{AP}    & $78.3_{\pm1.4}$ &   $84.3_{\pm0.5}$    & $84.7_{\pm0.3}$ & $\textbf{86.7}_{\pm0.5}$ \\
          & \textbf{FT}    & $76.7_{\pm2.2}$ &   $83.5_{\pm0.9}$    & $84.2_{\pm0.1}$ & $\textbf{87.3}_{\pm0.4}$ \\
    \midrule
    \multirow{2}[2]{*}{$\Delta_{3}^{\mathcal{T}_3} \!\rightarrow\! \Delta_{4}^{\mathcal{T}_3}$} & \textbf{AP}    & $48.2_{\pm2.9}$ &   $66.7_{\pm0.9}$    & $67.4_{\pm0.7}$ & $\textbf{68.3}_{\pm2.6}$ \\
          & \textbf{FT}    & $51.8_{\pm4.2}$ &   $62.8_{\pm3.0}$    & $65.2_{\pm1.4}$ & $\textbf{69.8}_{\pm1.8}$ \\
    \bottomrule
    \end{tabular}%
  \caption{Experiments on $\text{RoBERTa}_\texttt{BASE}$ when teacher models are adapted with the full-size dataset. Other settings are kept the same with Table~\ref{tab:fewshot_old_task} setting (a).}
  \label{tab:roberta_base_fulldata}%
\end{table}%

\subsection{Effects of Teacher Model Capability for Distillation-based Recyclable Tuning}

For experiments of setting (a) in distillation-based recyclable tuning (\cref{sec:application_distil}), the teacher model is trained with the same $32$-shot dataset as the student model. Here we explore whether a teacher model with stronger capabilities would conduce to the student's performance. Specifically, keeping all the other settings the same, we change the teacher model's data to the full-data size. The new results are placed in Table~\ref{tab:roberta_base_fulldata}, from which we conclude that: (1) our methods ($\mathcal{L}_\text{final}$, $\mathcal{L}_\text{ITP}$, and $\mathcal{L}_\text{final}$+\textit{Init.}) still outperform the baseline without knowledge distillation ($\mathcal{L}_\text{final}$-$\mathcal{L}_\text{KD}$); (2) comparing the student's performance in Table~\ref{tab:roberta_base_fulldata} and Table~\ref{tab:fewshot_old_task} setting (a), we find through learning from a more powerful teacher, the student's performance is improved as well.

\subsection{Experiments on Non-adjacent PLMs}
\label{sec:non_adjacent}

\begin{table}[!tbp]
  \centering
  \small
    \begin{tabular}{c@{~}c@{~~}c@{~}c@{~}c@{~}c@{~}}
    \toprule
    \multicolumn{2}{c}{\textbf{Method}} & $\mathcal{L}_\text{final}$-$\mathcal{L}_\text{KD}$  & $\mathcal{L}_\text{final}$ &  $\mathcal{L}_\text{ITP}$ & $\mathcal{L}_\text{final}$+\textit{Init.} \\
    \midrule
    \multicolumn{6}{c}{\textit{Few-shot teacher}} \\
    \midrule
    \multirow{2}[2]{*}{$\Delta_{1}^{\mathcal{T}_1} \!\rightarrow\! \Delta_{3}^{\mathcal{T}_1}$} & \textbf{AP}    & $60.5_{\pm2.1}$ &   $66.3_{\pm1.5}$    & $\textbf{67.5}_{\pm1.7}$ & $67.2_{\pm1.5}$ \\
          & \textbf{FT}    & $61.9_{\pm1.3}$ &   $64.7_{\pm0.9}$    & $64.8_{\pm0.8}$ & $\textbf{65.4}_{\pm1.3}$ \\
    \midrule
    \multirow{2}[2]{*}{$\Delta_{1}^{\mathcal{T}_1} \!\rightarrow\! \Delta_{4}^{\mathcal{T}_1}$} & \textbf{AP}    & $56.6_{\pm1.1}$ &  $57.9_{\pm1.5}$     & $\textbf{65.3}_{\pm1.7}$ & $64.9_{\pm3.6}$ \\
          & \textbf{FT}    & $59.7_{\pm2.3}$ &   $62.9_{\pm2.4}$    & $64.4_{\pm0.5}$ & $\textbf{65.1}_{\pm2.2}$ \\
    \midrule
    \multicolumn{6}{c}{\textit{Full-data teacher}} \\
    \midrule
    \multirow{2}[2]{*}{$\Delta_{1}^{\mathcal{T}_1} \!\rightarrow\! \Delta_{3}^{\mathcal{T}_1}$} & \textbf{AP}    & $60.5_{\pm2.1}$ &   $74.2_{\pm0.9}$    & $77.8_{\pm0.6}$ & $\textbf{78.0}_{\pm0.5}$ \\
          & \textbf{FT}    & $61.9_{\pm1.3}$ &   $70.9_{\pm0.5}$    & $73.3_{\pm1.0}$ & $\textbf{77.3}_{\pm0.7}$ \\
    \midrule
    \multirow{2}[2]{*}{$\Delta_{1}^{\mathcal{T}_1} \!\rightarrow\! \Delta_{4}^{\mathcal{T}_1}$} & \textbf{AP}    & $56.6_{\pm1.1}$ &   $68.6_{\pm0.7}$    & $75.6_{\pm0.7}$ & $\textbf{76.7}_{\pm0.1}$ \\
          & \textbf{FT}    & $59.7_{\pm2.3}$ &  $69.8_{\pm0.5}$     & $74.0_{\pm0.5}$ & $\textbf{76.0}_{\pm0.8}$ \\
    \bottomrule
    \end{tabular}%
  \caption{Experiments for distillation-based recyclable tuning between non-adjacent PLMs, i.e., ($\mathcal{M}_1$, $\mathcal{M}_3$) and ($\mathcal{M}_1$, $\mathcal{M}_4$). We follow the setting (a) in \cref{sec:application_distil}. The teacher model is trained using either the $32$-shot data or the full data. The student model is trained using the $32$-shot data.}
  \label{tab:non_adjacent_exp}%
\end{table}%

For most of the experiments, we mainly focus on recyclable tuning between adjacent PLMs. We contend that the proposed methods should also work for non-adjacent PLMs since they are still guaranteed with close connections. To demonstrate this, we take the distillation-based recyclable tuning as an example. Specifically, we evaluate the distillation-based recyclable tuning between ($\mathcal{M}_1$, $\mathcal{M}_3$) and ($\mathcal{M}_1$, $\mathcal{M}_4$) using $\mathcal{T}_1$, and largely follow the settings in \cref{sec:application_distil}. We choose setting (a) in \cref{sec:application_distil}, and the only difference is that the teacher model $\mathcal{M}_1$ is trained either using the $32$-shot dataset (dubbed as \textit{few-shot teacher}) or the full dataset (dubbed as \textit{full-data teacher}). While the student model is trained using the $32$-shot dataset. In this way, we could understand the role of the teacher model in knowledge distillation.

The results are placed in Table~\ref{tab:non_adjacent_exp}, from which we find that: (1) introducing knowledge distillation ($\mathcal{L}_\text{final}$) improves the performance than only using task loss ($\mathcal{L}_\text{final}$-$\mathcal{L}_\text{KD}$) and (2) introducing the parametric knowledge either through interpolation distillation ($\mathcal{L}_\text{ITP}$) or weight initialization ($\mathcal{L}_\text{final}$+\textit{Init.}) could further improve the task performance. Both conclusions are aligned with those obtained on adjacent PLMs. This demonstrates our claim that our recyclable tuning is not limited to adjacent PLMs, but also non-adjacent ones. Finally, we observe that the student performance when the teacher is trained using full data is much better, which shows the benefits of learning from a more advanced teacher.

\subsection{Distillation-based Recyclable Tuning Experiments using $\text{RoBERTa}_\texttt{LARGE}$}
\label{sec:roberta_large_distil}

\begin{table}[!tbp]
  \centering
  \small
    \begin{tabular}{c@{~}c@{~~}c@{~}c@{~}c@{~}c@{~}}
    \toprule
    \multicolumn{2}{c}{\textbf{Method}} & $\mathcal{L}_\text{final}$-$\mathcal{L}_\text{KD}$  & $\mathcal{L}_\text{final}$ &  $\mathcal{L}_\text{ITP}$ & $\mathcal{L}_\text{final}$+\textit{Init.} \\
    \midrule
    \multicolumn{6}{c}{\textit{Few-shot teacher}} \\
    \midrule
    \multirow{2}[2]{*}{$\Delta_{1}^{\mathcal{T}_1} \!\rightarrow\! \Delta_{2}^{\mathcal{T}_1}$} & \textbf{AP}    & $64.6_{\pm1.2}$ &   $\textbf{69.3}_{\pm0.8}$    & $70.2_{\pm0.1}$ & $69.2_{\pm1.1}$ \\
          & \textbf{FT}    & $64.7_{\pm2.7}$ &   $70.3_{\pm1.6}$    & $70.9_{\pm2.0}$ & $\textbf{72.5}_{\pm1.1}$ \\
    \midrule
    \multicolumn{6}{c}{\textit{Full-data teacher}} \\
    \midrule
    \multirow{2}[2]{*}{$\Delta_{1}^{\mathcal{T}_1} \!\rightarrow\! \Delta_{2}^{\mathcal{T}_1}$} & \textbf{AP}    & $64.6_{\pm1.2}$ &   $78.8_{\pm0.6}$    & $\textbf{82.8}_{\pm0.3}$ & $82.4_{\pm0.5}$ \\
          & \textbf{FT}    & $64.7_{\pm2.7}$ &   $76.9_{\pm0.5}$    & $79.8_{\pm1.0}$ & $\textbf{82.1}_{\pm0.3}$ \\
    \bottomrule
    \end{tabular}%
  \caption{Experiments of distillation-based recyclable tuning for $\text{RoBERTa}_\texttt{LARGE}$ ($\mathcal{M}_1$, $\mathcal{M}_2$). We follow the setting (a) in \cref{sec:application_distil}. The teacher model is trained using either the $32$-shot data or the full data. The student model is trained using the $32$-shot data.}
  \label{tab:roberta_large}%
\end{table}%

Previous experiments for distillation-based recyclable tuning are based on $\text{RoBERTa}_\texttt{BASE}$, now we turn to $\text{RoBERTa}_\texttt{LARGE}$ to show that our proposed methods are model-agnostic. We experiment with $\mathcal{M}_1$ and $\mathcal{M}_2$ using the task \textsc{ChemProt}. Other settings are kept the same as those in \cref{sec:non_adjacent}. In Table~\ref{tab:roberta_large}, we show that the results are generally aligned with our conclusions before. These results also reflect that our proposed method is agnostic to the specific PLM chosen.

\subsection{Effects of Data Size for Distillation-based Recyclable Tuning}

\begin{figure}[!t]
    \centering
    \subfigure{\includegraphics[width=0.235\textwidth]{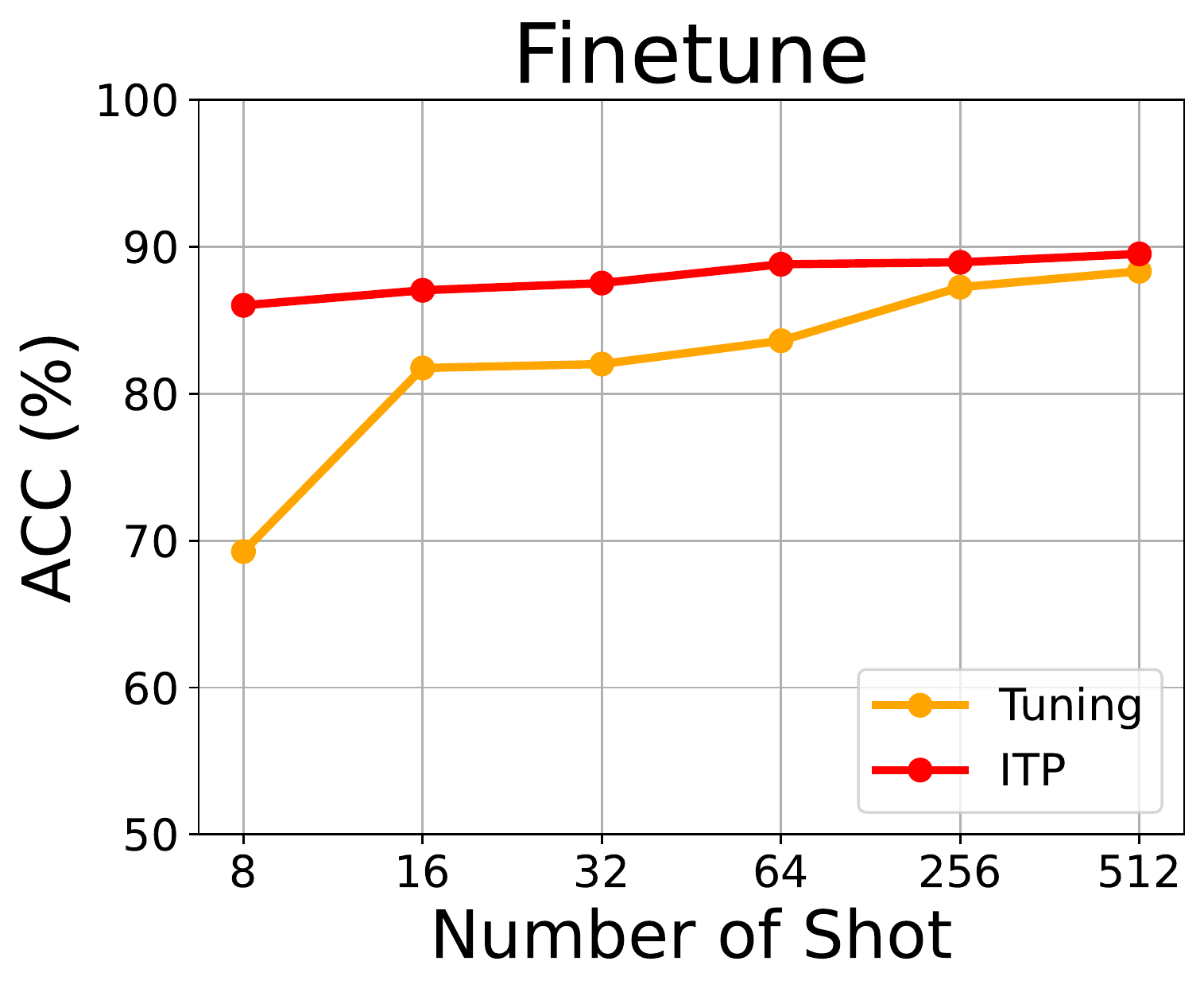}}     \subfigure{\includegraphics[width=0.235\textwidth]{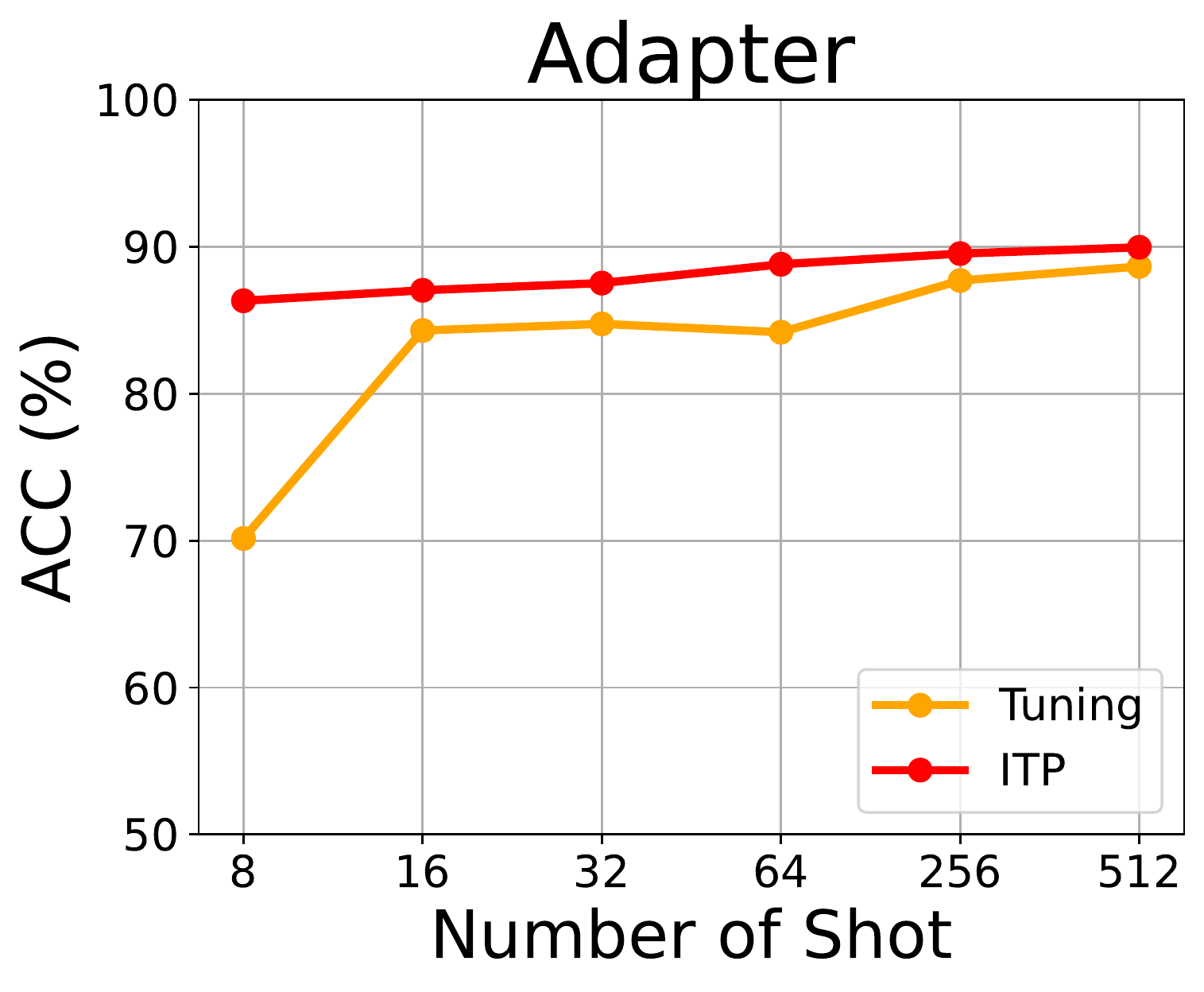}} 
    \caption{Performance variation of distillation-based recyclable tuning on IMDB at different data scales. We compare two methods: the conventional tuning method (Tuning), and our interpolation distillation method (ITP).}
    \label{fig:data_size}
\end{figure}

Taking a step further, we study the performance of our distillation-based recyclable tuning at different data scales. Specifically, we focus on $\mathcal{T}_2$ (IMDB) for recycling $\mathcal{M}_1$'s outdated weights to $\mathcal{M}_2$, where $\mathcal{M}_1$ is adapted using the full dataset, and $\mathcal{M}_2$ is trained with $\{8, 16, 32, 64\}$-shot dataset, respectively. By comparing the method mentioned in \cref{sec:interpolation_distillation} ($\mathcal{L}_\text{ITP}$) with only the task loss $\mathcal{L}_{\mathcal{T}}$, we visualize the performance variation in Figure~\ref{fig:data_size}, from which we observe that: $\mathcal{L}_\text{ITP}$ surpasses only the task loss ($\mathcal{L}_{\mathcal{T}}$) in general. However, with the data scale increasing, the improvement becomes smaller. This is because $\mathcal{M}_2$ is more adept at $\mathcal{T}_2$ than $\mathcal{M}_1$ due to the incremental knowledge acquisition of $\mathcal{D}_2$. When there are only a few examples to train $\mathcal{M}_2$, the teacher model has the advantage of more labeled data. However, with the data size of the student gradually approaching that of the teacher, learning from the teacher gradually becomes redundant. The student model could well master the downstream knowledge on its own.

\section{Training Details}
\label{sec:all_training_detail}
We ensure that all the artifacts used in this paper are consistent with their intended use.

\begin{table}
  \centering
  \small
    \begin{tabular}{c@{~~~}c@{~~~}c@{~~~}c@{~~~}c}
    \toprule
    \textbf{Task} & \textbf{LR} & \textbf{BS} & \textbf{S/E(AP)} & \textbf{S/E(FT)} \\
    \midrule
    \textsc{ChemProt} & $2\times10^{-5}$ & $16$ & $25$ epochs & $10$ epochs \\
    \textsc{IMDB} & $2\times10^{-5}$ & $16$ & $25$ epochs & $10$ epochs \\
    \textsc{ACL-ARC} & $2\times10^{-5}$ & $16$ & $25$ epochs & $10$ epochs \\
    MNLI & $2\times10^{-5}$ & $32$ & $50$k steps & $50$k steps \\
    ANLI & $2\times10^{-5}$ & $32$ & $50$k steps & $50$k steps \\
    SICK & $2\times10^{-5}$ & $32$ & $50$k steps & $50$k steps \\
    R. Tomatoes & $2\times10^{-5}$ & $32$ & $50$k steps & $50$k steps \\
    A. Polarity & $2\times10^{-5}$ & $16$ & $15$k steps & $15$k steps \\
    SST-2 & $2\times10^{-5}$ & $16$ & $15$k steps & $15$k steps \\
    H. Speech & $2\times10^{-5}$ & $16$ & $15$k steps & $15$k steps \\
    T. Hate & $2\times10^{-5}$ & $16$ & $15$k steps & $15$k steps \\
    T. Offensive & $2\times10^{-5}$ & $16$ & $15$k steps & $15$k steps \\
    \bottomrule
    \end{tabular}%
  \caption{Hyper-parameters used for downstream adaptation (LR: learning rate, BS: batch size, S / E: maximum step / maximum epoch. AP and FT refer to adapter tuning and fine-tuning.}
  \label{tab:hyperparameter_empirical_analysis}%
\end{table}%

\subsection{Pre-training}
\label{sec:training_detail_lifelong_pretrain}
We conduct pre-training using $8$ NVIDIA V100 GPUs based on \texttt{fairseq}\footnote{\url{https://github.com/pytorch/fairseq}}~\cite{ott2019fairseq}. We choose Adam~\citep{kingma2014adam} as the optimizer. The hyper-parameters ($\epsilon, \beta_1, \beta_2$) for Adam are set to $1\times10^{-6}, 0.9, 0.98$, respectively. The dropout rate and weight decay are set to $0.1$ and $0.01$, respectively. The total number of parameters of $\text{RoBERTa}_\texttt{BASE}$ and $\text{RoBERTa}_\texttt{LARGE}$ are $125$M and $355$M, respectively. We implement pre-training using the codes of \citet{qin-etal-2022-knowledge}.

\paragraph{Continual Pre-training.}
We start with the official $\text{RoBERTa}$ model and sequentially pre-train the PLM on $4$ domains. For each domain, we set the batch size to $2048$, the training steps to $12.5$k, and the max sequence length to $512$.

\paragraph{Pre-training from Scratch.}
For $\mathcal{M}_\text{IND}$ that is pre-trained from scratch, we follow the model structure of $\text{RoBERTa}_\texttt{BASE}$, and pre-train the model on the concatenation of Wikipedia and BookCorpus~\citep{zhu2015aligning}, which is the same as the pre-training corpus of BERT~\citep{devlin2018bert}. We pre-train the model for $125$k steps, using a batch size of $2048$ and a sequence length of $512$. The total computations involved are roughly comparable to those of $\text{BERT}_\texttt{BASE}$. $\mathcal{M}_\text{IND}$ has totally different initialization and pre-training corpus than the official $\text{RoBERTa}_\texttt{BASE}$, which helps us understand the property between independently trained PLMs.

\subsection{Empirical Analyses}
\label{sec:training_detail_empirical}

\paragraph{Model Compatibility Analysis.}
We adapt the initial PLM $\mathcal{M}_{0}$ on two tasks \textsc{ChemProt} and MNLI. The training hyper-parameters conform to those listed in Table~\ref{tab:hyperparameter_empirical_analysis}. All experiments are conducted $3$ times with different random seeds, and we report the average results.

\paragraph{Linear Mode Connectivity Analysis.}
All the training hyper-parameters conform to those in Table~\ref{tab:hyperparameter_empirical_analysis}. The endpoints are adapted three times using different random seeds. We test the performance of $25$ evenly distributed points along the linear path and two endpoints. We report the average performance over three random seeds.

\begin{table}[!t]
  \centering
  \small
    \begin{tabular}{l|ccc}
    \toprule
    \textbf{Target Task} & $\mathcal{T}_{\text{diff}}$  & $\mathcal{T}_{\text{sim}}$ & EI (steps) \\
    \midrule
    ANLI  & SST-2  & MNLI & $1000$ \\
    SICK  & SST-2  & MNLI & $50$\\
    SST-2 & MNLI  & A. Polarity & $60$ \\
    R. Tomatoes & MNLI  & A. Polarity & $100$ \\
    H. Speech & MNLI  & T. Hate & $300$ \\
    T. Offensive & MNLI  & T. Hate & $40$ \\
    \bottomrule
    \end{tabular}%
  \caption{The selection of source tasks and target tasks for experiments in \cref{sec:application_init}. For each target task, we list both $\mathcal{T}_\text{diff}$ and $\mathcal{T}_\text{sim}$. We also report the evaluation interval (EI) w.r.t. training steps for each of the $6$ target tasks.}
  \label{tab:convergence_task}%
\end{table}

\paragraph{Functional Similarity Analysis.}
We adapt different PLMs on task \textsc{ChemProt} using the hyper-parameters listed in Table~\ref{tab:hyperparameter_empirical_analysis}. We randomly sample one instance\footnote{We find empirically that the results and conclusions are very consistent across different random samples.} from \textsc{ChemProt} and feed it into different PLMs to obtain the scores after the self-attention computation. We draw the attention scores for the first $25$ tokens of the sampled instance.

\subsection{Methods and Experiments}
\label{sec:training_detail_application}
For the optimizer of all the experiments in \cref{sec:application}, we choose AdamW~\citep{loshchilov2017decoupled}.

\paragraph{Initialization-based Recyclable Tuning.}
We adapt $\mathcal{M}_{0}$ on the source tasks using the hyper-parameters listed in Table~\ref{tab:hyperparameter_empirical_analysis}. The adapted weights are further used as target tasks' initialization (except the \textit{Random} setting). The target tasks' training configurations also conform to Table~\ref{tab:hyperparameter_empirical_analysis}. We conduct the experiments for $3$ times with different random seeds and report the average performance. The choices of $\mathcal{T}_\text{diff}$ and $\mathcal{T}_\text{sim}$ for different target tasks are shown in Table~\ref{tab:convergence_task}. The evaluation interval for each target task is also reported in Table~\ref{tab:convergence_task}.

\paragraph{Distillation-based Recyclable Tuning.}

\begin{figure}[!t]
    \centering
    \subfigure{\includegraphics[width=0.48\textwidth]{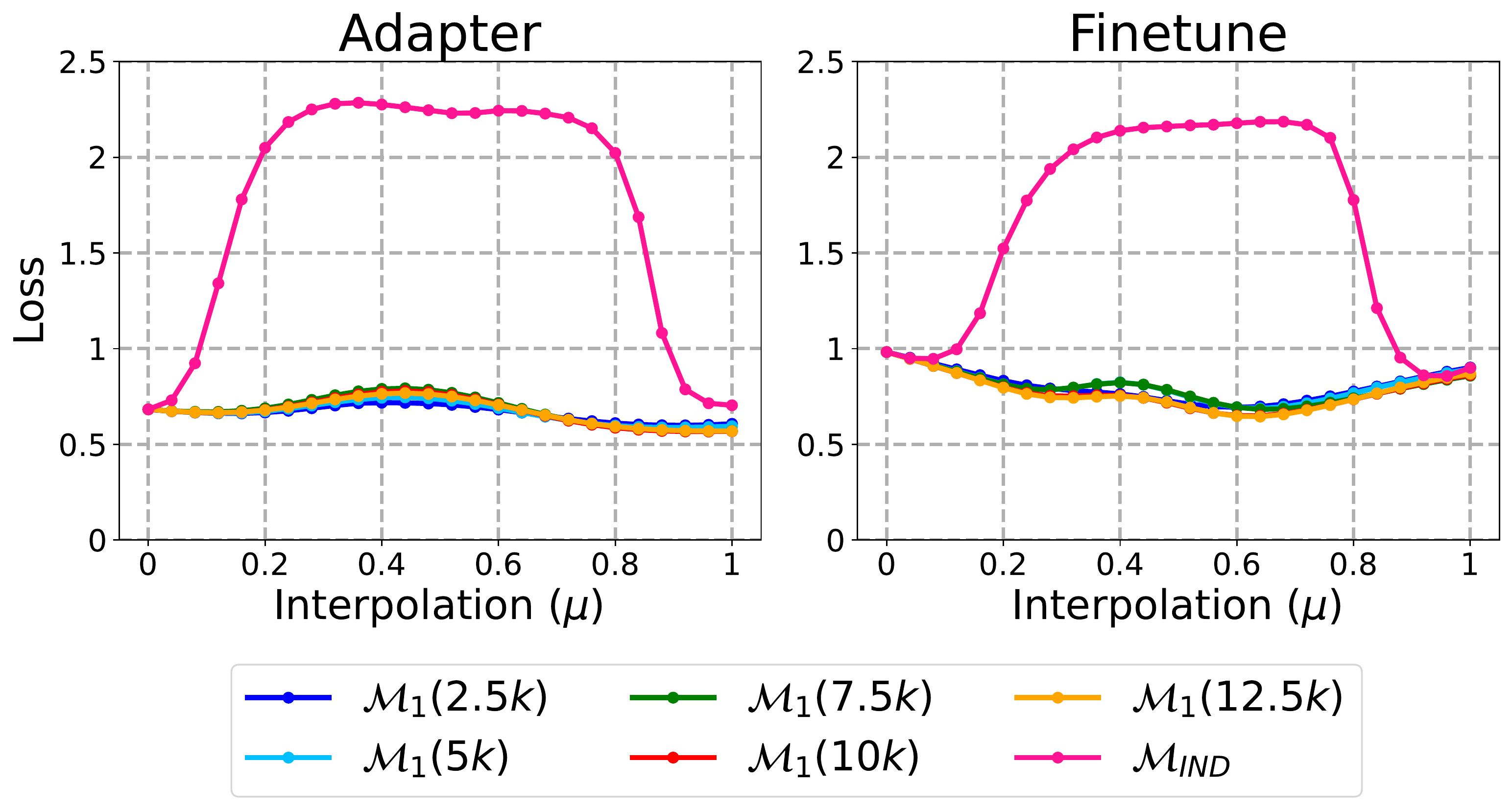}} 
    \caption{The loss of linear interpolations between two adapted PLMs ($\theta_{0}^\mathcal{T}$ and $\theta_{1}^\mathcal{T}(t)$) on \textsc{ChemProt}. The corresponding visualization for performance is shown in Figure~\ref{fig:mode_connectivity}.}
    \label{fig:mode_connectivity_loss}
\end{figure}

\begin{figure}[!t]
    \centering
    \subfigure{\includegraphics[width=0.48\textwidth]{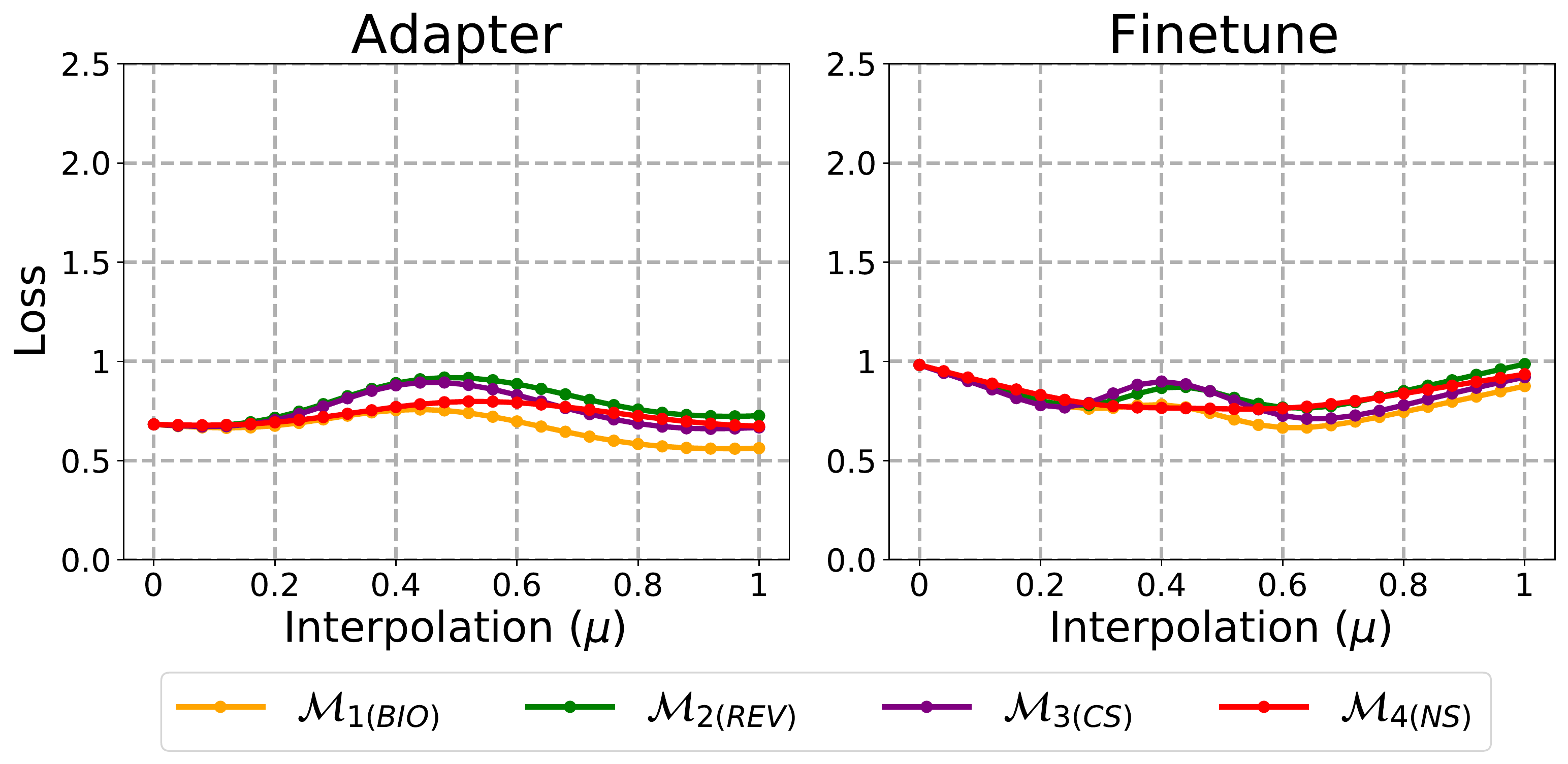}} 
    \caption{Mode connectivity (loss evaluation) between continually pre-trained PLMs across $4$ domains and the initial $\mathcal{M}_0$. The corresponding visualization for performance is shown in Figure~\ref{fig:mode_connectivity_multiple_domain}.}
    \label{fig:mode_connectivity_multiple_domain_loss}
\end{figure}

We set the maximum training step for \textsc{ChemProt} and \textsc{ACL-ARC} to $100$k and the maximum training step for \textsc{IMDB} to $50$k. The learning rate and batch size are set to $1\times10^{-4}$ and $2$, respectively. We warm up the learning rate for the first $8\%$ percentage of total training steps. We report the average results over $3$ different random seeds. As for other hyper-parameters discussed in \cref{sec:application_distil}, we perform grid search for $\beta$ over \{$0.1$, $0.3$\}, and $\alpha(1-\beta)$ over \{$0$, $1$, $5$, $10$, $50$, $100$\}. We also conduct a grid search for the temperature in knowledge distillation loss over \{$10$, $20$\} when calculating $\textrm{KL}(\mathcal{P}(x, \mathcal{M}_i) || \mathcal{P}(x, \mathcal{M}_{i+1}))$. We select the best-performing combination of these hyper-parameters and then report the performance. Our grid search is performed for our method and all the baseline methods for a fair comparison.

\section{The Visualization of Loss for Linear Mode Connectivity Analysis}
\label{sec:visualization_loss}
When conducting experiments for the mode connectivity analysis in the main paper, we mainly resort to performance as the evaluation protocol for the interpolations following \citet{qin2022exploring}. In this section, we show the corresponding visualization of loss for Figure~\ref{fig:mode_connectivity} and Figure~\ref{fig:mode_connectivity_multiple_domain}, see Figure~\ref{fig:mode_connectivity_loss} and Figure~\ref{fig:mode_connectivity_multiple_domain_loss}. From these figures, we conclude that a significant loss barrier generally indicates the existence of a large performance drop.

\section{Comparison of Initialization-based and Distillation-based Recyclable Tuning}
\label{sec:compare_init_distil}

Both initialization-based and distillation-based methods serve as powerful ways for recyclable tuning under the continual pre-training scenario. Both methods have their own advantages, where the initialization-based method can bring faster convergence and performance improvement, while the distillation-based method can bring improvement in performance as well (but may be less efficient). In addition, both methods can be combined with each other to further improve performance.

In terms of practical application scenarios, both methods are slightly different. For one thing, the initialization-based method requires that the architectures of the new PLM and the old PLM are the same. This requirement may be infeasible for broader application scenarios, such as recyclable tuning between different PLMs as discussed in \cref{sec:discussion}. For another, the initialization-based method typically requires access to the parameters of the outdated adapted weights. This can be a practical issue due to model privacy concerns. While some customers are willing to share their adapted weights on public platforms like AdapterHub~\citep{pfeiffer2020AdapterHub}, a majority of adapted weights are publicly unavailable. In contrast, the distillation-based method can be achieved without access to the model weights, but through receiving model inference from the owner (e.g., API-based online knowledge transfer~\citep{krishna2019thieves}). In this sense, the distillation-based method could protect the model privacy to a certain degree. 

\section*{Broader Impacts}

This research has the potential to have a broad impact in several ways.

\begin{itemize} [topsep=1pt, partopsep=1pt, leftmargin=12pt, itemsep=-3pt]
    \item First, recyclable tuning could improve the efficiency of adapting PLMs to new tasks. By recycling adapted weights from previous tasks, the need for costly retraining can be reduced, potentially making it more feasible to apply PLMs in a wider range of scenarios.
    \item Second, the results of this research could have implications for the sustainability of machine learning systems. Reusing adapted weights rather than discarding them can help us reduce the carbon footprint and resource consumption of PLM adaptation, making it more environmentally friendly.
    \item Third, this research has the potential to benefit a wide range of stakeholders, including researchers, developers, and users of PLMs. Researchers can use the proposed task and benchmark to develop and evaluate new techniques for recyclable tuning, while developers can apply these techniques to improve the efficiency and sustainability of PLM-based systems. Finally, users of PLMs can benefit from the reduced costs and improved performance made possible by recyclable tuning.
\end{itemize}

Overall, this research on recyclable tuning for continual pre-training has the potential to have a wide-ranging impact on the efficiency, sustainability, and practicality of machine learning systems.

\end{document}